\documentclass[runningheads]{llncs}

\usepackage{eccv}

\usepackage{eccvabbrv}

\usepackage{graphicx}
\usepackage{booktabs}

\usepackage{multirow}
\usepackage{bm}
\usepackage{wrapfig}
\usepackage{amssymb}
\usepackage{amsmath}
\usepackage{mathrsfs}
\usepackage{makecell}
\usepackage{colortbl}
\usepackage{pifont} 

\usepackage{array}

\definecolor{GrassGreen}{RGB}{146,208,80}
\definecolor{SkyBlue}{RGB}{4,175,252}
\definecolor{FluoGreen}{RGB}{24,229,25}
\definecolor{Fig1aGreen}{RGB}{106, 234, 102}
\definecolor{Fig1aBlue}{RGB}{0, 176, 240}
\definecolor{EncoderBlue}{RGB}{166, 203, 220}
\definecolor{DecoderGreen}{RGB}{131, 186, 158}
\definecolor{Pending}{RGB}{255, 0, 0}
\definecolor{red}{RGB}{255, 0, 0}
\definecolor{green}{RGB}{0, 255, 0}
\definecolor{blue}{RGB}{0, 0, 255}
\definecolor{AblationRed}{RGB}{251, 221, 238}
\definecolor{AblationRedWord}{RGB}{231, 146, 238}
\definecolor{AblationGreen}{RGB}{193, 220, 206}
\definecolor{AblationGreenWord}{RGB}{160, 191, 118}
\definecolor{AblationOrangeWord}{RGB}{254, 123, 0}
\definecolor{red2}{RGB}{225, 0, 0}
\definecolor{blue3}{RGB}{0, 0, 150}
\definecolor{green3}{RGB}{0, 180, 0}
\definecolor{myblue}{rgb}{0.0, 0.3, 0.6}

\newcommand{\CircledNumber}[1]{\normalsize{\textcircled{\small{#1}}}}
\newcommand{\CircledNumberX}[1]{\normalsize{\textcircled{\scriptsize{#1}}}}

\definecolor{TabGray}{RGB}{232, 232, 232}

\usepackage{silence}
\definecolor{darkpastelgreen}{rgb}{0.01, 0.75, 0.24}
\definecolor{darkpink}{rgb}{0.91, 0.33, 0.5}
\newcommand{\more}[1]{$\textcolor{darkpink}{\scriptstyle \blacktriangle#1\%}$}
\newcommand{\less}[1]{$\textcolor{darkpastelgreen}{\scriptstyle \blacktriangledown#1\%}$}

\newcommand{\cm}{\checkmark}
\newcommand{\dg}{$^\dag$}

\usepackage[table]{xcolor}
\usepackage{pgf}
\newcommand{\colorednumber}[1]{%
    \pgfmathsetmacro{\myratio}{(#1-226)/300*100}%
    \pgfmathsetmacro{\safecl}{max(0, min(100, \myratio))}%
    \edef\temp{\noexpand\cellcolor{red!\safecl!green}}\temp
    \pgfmathparse{#1 > 450 ? "white" : "black"}%
    \color{\pgfmathresult}#1%
}

\newcommand{\colorednumberflops}[1]{%
    \pgfmathsetmacro{\myratio}{(#1-154)/(319-154)*100}%
    \pgfmathsetmacro{\safecl}{max(0, min(100, \myratio))}%
    \edef\temp{\noexpand\cellcolor{red!\safecl!green}}\temp
    \pgfmathparse{#1 > 277 ? "white" : "black"}%
    \color{\pgfmathresult}#1%
}

\usepackage{capt-of}

\usepackage[accsupp]{axessibility}  

\usepackage{hyperref}

\usepackage{orcidlink}

\usepackage{marvosym}
\usepackage{pifont}

\begin{document}
\title{Moebius: 0.2B Lightweight Image Inpainting Framework with 10B-Level Performance}
\titlerunning{~}
\author{
Kangsheng Duan\textsuperscript{1,$*$} \quad
Ziyang Xu\textsuperscript{1,$*$,\dag} \orcidlink{0009-0002-7682-6995} \quad
Wenyu Liu\textsuperscript{1} \orcidlink{0000-0002-4582-7488} \quad
Xiaohu Ruan\textsuperscript{2} \orcidlink{0009-0008-4975-2103}
\\
Xiaoxin Chen\textsuperscript{2} \orcidlink{0009-0008-5908-847X} \quad
Xinggang Wang\textsuperscript{1,\Letter} \orcidlink{0000-0001-6732-7823}
}
\authorrunning{~}
\institute{Huazhong University of Science and Technology \and VIVO AI Lab}
\maketitle

\makeatletter\def\Hy@Warning#1{}\makeatother
\let\thefootnote\relax\footnotetext{\textsuperscript{$*$} Equal contribution. Completed this work as interns at VIVO AI Lab.}
\let\thefootnote\relax\footnotetext{\textsuperscript{\dag} Project leader.}
\let\thefootnote\relax\footnotetext{\textsuperscript{\Letter} Corresponding author: Xinggang Wang <xgwang@hust.edu.cn>.}

\begin{abstract}
  While 10B-level industrial foundation models have pushed the boundaries of image inpainting, their prohibitive computational costs severely hinder practical deployment. Constructing a highly optimized task-specific specialist offers a promising solution; however, extreme structural compression inevitably triggers a severe representation bottleneck. To conquer this, we propose Moebius, a highly efficient lightweight inpainting framework. We systematically reconstruct the diffusion backbone by introducing the Local-$\lambda$ Mix Interaction ($L\lambda MI$) block. Comprising Local-$\lambda$ and Interactive-$\lambda$ modules, it elegantly summarizes spatial contexts and global semantic priors into fixed-size linear matrices, preserving complex latent interactions while drastically shedding parameters. Furthermore, to unlock the full representational capacity of this highly compact architecture, we synergistically pair it with an adaptive multi-granularity distillation strategy. Operating strictly within the latent space to avoid expensive pixel-space decoding, this strategy dynamically balances multiple gradient-based losses to achieve high-fidelity alignment. Extensive experiments across natural and portrait benchmarks demonstrate that this optimal synergy enables Moebius to rival or even surpass the generation quality of the 10B-level industrial generalist FLUX.1-Fill-Dev. Remarkably, Moebius achieves this using less than 2\% of the parameters (0.22B vs. 11.9B) while delivering a $>15\times$ acceleration in total inference time, setting a new efficiency standard for high-fidelity inpainting. Project page at https://hustvl.github.io/Moebius.
\end{abstract}

\section{Introduction}\label{sec:intro}
\vspace{-0.5em}
Image inpainting \cite{Bertalmio2000imageinpaint}, a fundamental task in computer vision aimed at reconstructing missing regions with visually coherent content, has been profoundly revolutionized by the rapid evolution of diffusion models \cite{xu2025pixelhacker,Rombach2022LDM,flux2024}. Recently, industrial-grade generalist foundation models, such as FLUX.1-Fill-Dev \cite{flux2024} and SD3.5 Large-Inpainting \cite{esser2024SD3}, have pushed the boundaries of zero-shot generation quality by scaling parameters to the 10-billion (10B) level. However, the exorbitant computational costs and massive memory footprints of these colossal models severely hinder their practical deployment, particularly on resource-constrained devices or in latency-sensitive applications. This dilemma motivates us to rethink the current scaling paradigm: Can a highly optimized, lightweight task-specific specialist bridge the massive scale gap and rival the performance of 10B-level generalists? While massive generalist models excel in zero-shot versatility, fine-tuning on established academic benchmarks (e.g., Places2 \cite{zhou2017places}, CelebA-HQ \cite{karras2018celebahq}) remains the standard evaluation paradigm in the inpainting community \cite{suvorov2021lama, li2022mat} to unlock and rigorously assess a model's capacity for specific restoration tasks. Following this well-established practice, we demonstrate that by pushing architectural efficiency to its limits, a 0.2B-parameter specialist model can successfully overcome the capacity bottleneck and match the high-fidelity generation capabilities of its 10B-level counterparts.

\vspace{-0.5em}
To construct such a highly efficient specialist, a natural progression is to compress existing diffusion architectures. Recent advancements like PixelHacker \cite{xu2025pixelhacker} have pioneered efficient high-fidelity inpainting by introducing the Latent Categories Guidance (LCG) paradigm and employing Gated Linear Attention (GLA) \cite{yang2024gla} to reduce computational overhead. Despite these algorithmic innovations, its backbone still comprises nearly one billion parameters, which remains prohibitive for edge deployment. A naive solution to further compress the model would be directly substituting its standard convolutions and attention blocks with off-the-shelf lightweight operators, such as Depthwise Convolutions (DWConv) \cite{Sandler2018MobileNetV2} and linear attention mechanisms \cite{yang2024gla,dao2022flashattention,dao2023flashattention2}. However, our empirical analysis reveals that such straightforward structural reductions inevitably trigger a severe representation bottleneck. In the intricate task of image inpainting, which demands rigorous semantic reasoning and precise spatial-texture alignment, these naive lightweight models suffer from a catastrophic degradation in generation quality. Furthermore, many efficient operators are architecturally constrained; for instance, while GLA \cite{yang2024gla} is highly efficient for self-attention, it inherently lacks the formulation to perform the cross-attention operations essential for integrating external semantic priors like LCG \cite{xu2025pixelhacker}.

\vspace{-0.5em}
To conquer this representation bottleneck and achieve an optimal balance between efficiency and quality, we propose Moebius, a highly efficient lightweight inpainting framework. We systematically reconstruct the foundational architecture through a rigorous synergy of structural design and knowledge distillation. First, to address the architectural constraints of existing linear attention, we introduce the Local-$\lambda$ and Interactive-$\lambda$ modules. By elegantly summarizing local spatial contexts and global semantic priors (e.g., LCG embeddings) into fixed-size linear matrices, these modules enable the network to perform both self- and cross-attention equivalents efficiently, preserving complex latent interactions while drastically shedding parameters. To push for extreme compactness, we further integrate highly compressed operators, such as DWConv and Mix-FFN \cite{xie2024sana, xie2025sana}, which collectively form our Local-$\lambda$ Mix Interaction ($L\lambda MI$) block. However, such extreme structural compression inherently risks weakening the network's representational capacity. To bridge this capacity gap without reintroducing architectural overhead, we propose an adaptive multi-granularity distillation strategy. By dynamically balancing multiple gradient-based losses, this strategy enables high-fidelity latent alignment between the lightweight specialist and a high-capacity teacher. Crucially, this distillation strategy exhibits a profound synergy with our architectural modifications—it effectively compensates for the representational drop incurred by extreme structural compression, unlocking the full potential of the $L\lambda MI$ blocks and culminating in the optimal, meticulously balanced architecture of Moebius.

\begin{table}[t]
\centering
\caption{\textbf{Breaking the impossible triangle of low-parameters, fast inference, and high generation quality.} Moebius achieves the lowest latency, FLOPs, and parameters, with remarkable generation quality. Note that industrial models require more steps (50/28 vs 20; default setting), making Moebius \textbf{15$\times$ faster} in total inference.}
\label{tab:cover}
\vspace{-0.8em}
\resizebox{1.0\linewidth}{!}{
\begin{tabular}{l|cc|cc|cc|ccccc}
\toprule
\multirow{2}{*}{\textbf{Model}}
& \multicolumn{2}{c|}{\textbf{Places2 (Small)}}  
& \multicolumn{2}{c|}{\textbf{CelebA-HQ (512)}} 
& \multicolumn{2}{c|}{\textbf{FFHQ (256)}}
& \textbf{Param.} 
& \multirow{2}{*}{\textbf{TFLOPs} $\downarrow$} 
& \textbf{Latency}
& \multirow{2}{*}{\textbf{Steps}}
& \textbf{Total} \\
&       $\mathbf{FID}\downarrow$    &     $\mathbf{LPIPS}\downarrow$  
&       $\mathbf{FID}\downarrow$    &     $\mathbf{LPIPS}\downarrow$  
&       $\mathbf{FID}\downarrow$    &     $\mathbf{LPIPS}\downarrow$  
& $\mathbf{(\times 10^9)}\downarrow$ 
& 
& $\mathbf{(ms/step)}\downarrow$
& 
& \textbf{Time(s)} $\downarrow$\\
\midrule
\rowcolor{AblationGreen}\textbf{Moebius} & 0.92 & 0.091 & 5.39 & 0.122 & 8.15 & 0.231 & 0.226 & 0.154 & 26.01 & 20 & 0.52 \\
\textbf{PixelHacker \cite{xu2025pixelhacker}} & 0.82\less{11} & 0.088\less{3} & 4.75\less{12} & 0.115\less{6} & 6.35\less{22} & 0.229\less{1} & 0.862\more{281}  & 0.338\more{119}  & 46.89\more{80}   & 20   & 0.94\more{81} \\
\textbf{SD3.5 Large-Inp. \cite{esser2024SD3}} & 3.02\more{228} & 0.105\more{15} & 11.80\more{119} & 0.134\more{10} & 109.42\more{1243} & 0.402\more{74} & 8.057\more{3465} & 8.657\more{5521} & 151.02\more{481} & 28\more{40}  & 4.23\more{713} \\
\textbf{FLUX.1-Fill-Dev \cite{flux2024}} & 0.94\more{2} & 0.099\more{9} & 10.13\more{88} & 0.141\more{16} & 11.19\more{37} & 0.268\more{16} & 11.902\more{5166}& 9.927\more{6346} & 161.01\more{519} & 50\more{150} & 8.05\more{1448} \\
\bottomrule
\end{tabular}
}
\vspace{-1.6em}
\end{table}

\vspace{-0.01em}
We conduct extensive experiments across natural (Places2 \cite{zhou2017places}) and portrait (CelebA-HQ \cite{karras2018celebahq}, FFHQ \cite{karras2018stylegan_ffhq}) benchmarks to rigorously validate the effectiveness of Moebius. As highlighted in Tab.~\ref{tab:cover}, Moebius achieves an outstanding inference latency of 26.01 ms/step, with theoretical FLOPs of 0.154 TFLOPs. When factoring in the required sampling steps, this translates to a remarkable $>15\times$ speed-up in total inference time compared to the 10B-level industrial SOTA, FLUX.1-Fill-Dev. Despite using less than 2\% of the parameters (0.22B vs. 11.9B), Moebius delivers comparable or even superior generation quality, demonstrating extreme architectural efficiency.

\vspace{-0.01em}
In summary, our main contributions are as follows:
\vspace{-0.4em}
\begin{itemize}
    \item We propose Moebius, a highly efficient lightweight image inpainting framework that sets a new efficiency standard. By functioning as a highly optimized task-specific specialist, it successfully bridges the massive scale gap, matching the performance of 10B-level generalist foundation models with only 0.22B parameters.
    \item We systematically conquer the representation bottleneck of compact networks by introducing the $L\lambda MI$ block alongside an adaptive multi-granularity distillation strategy. The optimal synergy between these structural and optimization designs preserves complex semantic reasoning capabilities while pushing compression to its limits.
    \item We provide rigorous empirical validation, including efficiency profiling and comprehensive blind user studies. Extensive evaluations across natural and portrait benchmarks, as well as real-world object removal scenarios, demonstrate Moebius’s superior performance-parameter-latency trade-off.
\end{itemize}

\vspace{-1.2em}
\section{Related Work}\label{sec:related}
\vspace{-0.5em}
\subsection{Efficient and Lightweight Architectures}
\vspace{-0.5em}
Designing compact yet effective architectures has been a long-standing goal in computer vision \cite{qin2024mobilenetv4}. Classical lightweight designs aim to reduce computational complexity and parameter count while preserving representational capacity \cite{li2022efficientformerv2}.
Among these efforts, DWConv \cite{Sandler2018MobileNetV2, tan2019efficientnet} and group convolutions \cite{zhang2018shufflenet} are widely adopted for efficient local feature extraction by decoupling spatial and channel interactions.
Meanwhile, low-rank FFN designs \cite{vaswani2017attention, shazeer2020glu, xie2024sana, xue2024openmoe, cai2023efficientvit} and linear attention mechanisms \cite{yang2024fla, yang2024gla, dao2022flashattention,Xu2024MoSt-DSA,xu2025garamost,bello2021lambdanetworks} have been proposed to enhance efficiency in transformer-based architectures \cite{xie2024sana, Peebles2022DiT, esser2024SD3, Zhu2025DiG} while maintaining representational ability.
Despite their success, these approaches often face inherent trade-offs between compactness and perceptual quality \cite{yao2025vavae}.
In this work, Moebius overcomes these limitations by systematically conquering the representation bottleneck of compact networks. Instead of naive module substitution, we introduce the Local-$\lambda$ and Interactive-$\lambda$ modules to elegantly summarize spatial and semantic contexts into fixed-size linear matrices, culminating in the $L\lambda MI$ block that pushes extreme compression while preserving rigorous semantic reasoning.
\vspace{-0.5em}

\vspace{-0.5em}
\subsection{Knowledge Distillation of Diffusion Models}
\vspace{-0.5em}
Knowledge distillation (KD) serves as a fundamental paradigm to transfer knowledge from a large, high-capacity teacher model to a lightweight student \cite{li2014KLDNN}.
Classical KD techniques typically operate on three supervision objectives: soft labels \cite{hinton2015KD}, feature maps \cite{romero2015fitnet, chen2021reviewKD, wang2023semKD, tian2022repdistiller, Sargsyan2023migan}, and perceptual metrics \cite{simonyan2015vgg, zhang2018lpips, yim2017giftKD, kang2024diffusion2gan}.
Recently, diffusion-specific KD methods \cite{salimans2022progressivedistillation, lu2025sCM, song2023consistencymodels} have emerged, enabling students to approximate teacher denoising dynamics with fewer sampling steps while preserving generation quality.
Distinct from conventional timestep distillation that focuses on sampling acceleration, our work targets architectural capacity transfer for extreme compression. 
To achieve this, Moebius employs an adaptive multi-granularity distillation strategy strictly within the latent space. 
By balancing multiple gradient-based objectives dynamically, our strategy perfectly compensates for the capacity drop incurred by extreme structural compression. This allows Moebius to seamlessly inherit high-level semantic priors and fine-grained textural consistency from a massive teacher, forging an optimal synergy that unlocks the full potential of our lightweight specialist.

\begin{figure}[t]
\centering
\includegraphics[width=0.98\linewidth]{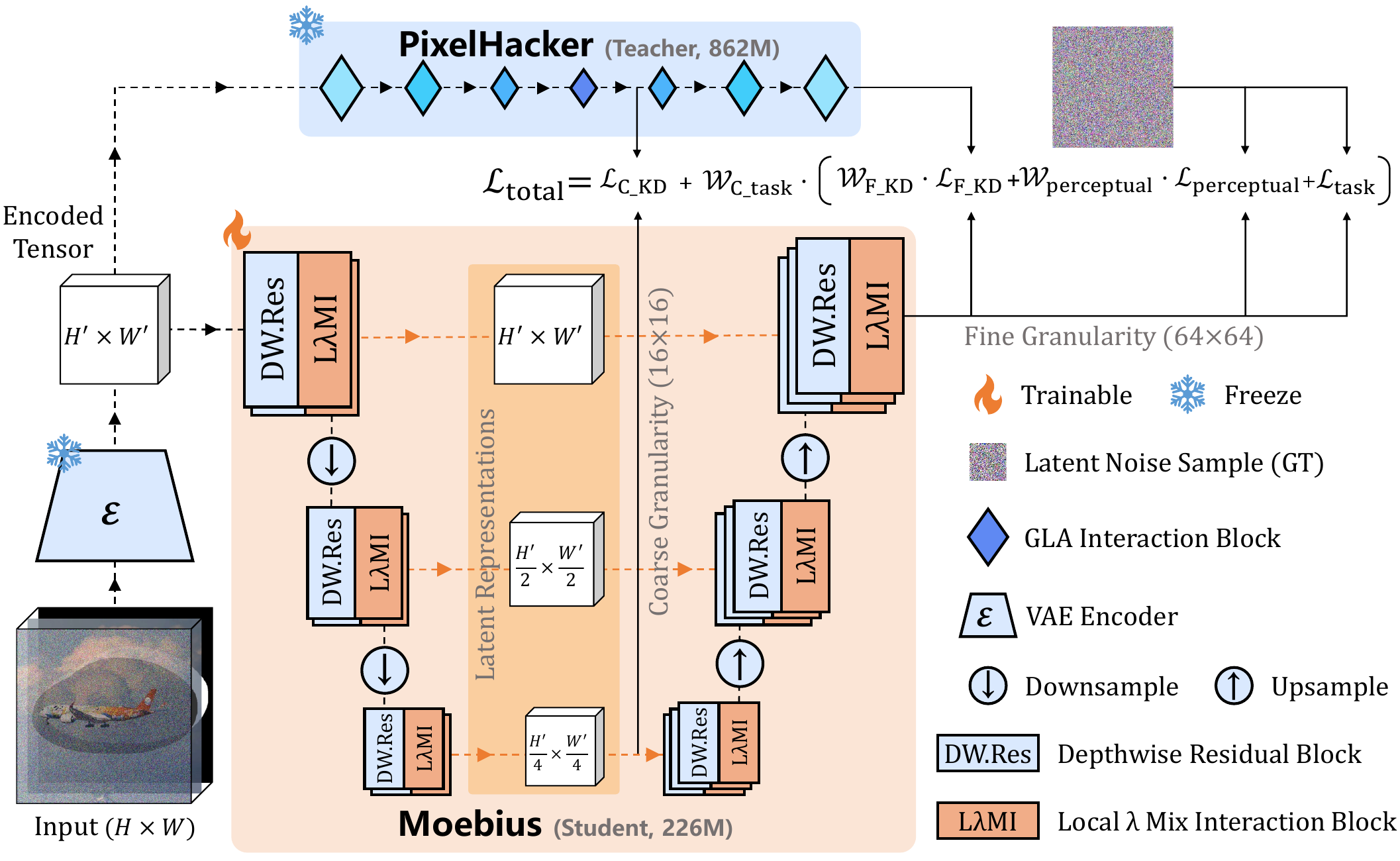}
\vspace{-0.5em}
\caption{
\textbf{Overall pipeline of Moebius.} We adopt the Latent Diffusion Model (LDM) \cite{Rombach2022LDM} framework equipped with Latent Categories Guidance (LCG) \cite{xu2025pixelhacker}. To achieve extreme architectural efficiency, the denoising U-Net is systematically restructured using our proposed $L\lambda MI$ blocks (detailed in Sec.~\ref{sec:architecture_evolution}). Furthermore, an adaptive multi-granularity distillation strategy (Sec.~\ref{sec:distillation}) is applied during training to align our lightweight specialist with the high-capacity teacher, successfully mitigating the capacity drop caused by extreme structural compression.
}
\label{fig:overall_pipeline}
\vspace{-1.7em}
\end{figure}

\vspace{-0.8em}
\section{Method}\label{sec:method}
\vspace{-0.5em}
In this section, we present the formulation of Moebius, a highly efficient light\-weight framework for image inpainting, as illustrated in Fig.~\ref{fig:overall_pipeline}. We first establish our baseline latent diffusion architecture in Sec.~\ref{sec:pipeline}. Subsequently, in Sec.~\ref{sec:architecture_evolution}, we present an empirical analysis identifying the representation bottleneck inherent in compact networks, and detail how we systematically resolve this by proposing the Local-$\lambda$ Mix Interaction ($L\lambda MI$) block. Finally, in Sec.~\ref{sec:distillation}, we introduce an adaptive multi-granularity distillation strategy designed to achieve optimal synergy with our lightweight architecture.

\vspace{-0.5em}
\subsection{Overall Pipeline}\label{sec:pipeline}
\vspace{-0.5em}
We adopt the Latent Diffusion Model (LDM) \cite{Rombach2022LDM} as our foundational framework. Let $x \in \mathbb{R}^{H \times W \times 3}$ denote an unmasked image, $m \in \{0,1\}^{H \times W}$ represent a binary mask indicating the missing regions, and $x_{m} = x \odot (1-m)$ denote the masked input image, where $\odot$ is the Hadamard product. Following the standard implementation of LDM for inpainting \cite{Rombach2022LDM,podell2023sdxl,esser2024SD3, ju2024brushnet}, both the clean image $x$ and the masked image $x_{m}$ are encoded into the latent space using a pre-trained VAE \cite{Rombach2022LDM}. Specifically, the masked image is encoded into $z_{m} = \mathcal{E}(x_{m})$ and concatenated with the downsampled mask to form the spatial reference $z_{s}$. Concurrently, the clean latent representation $z = \mathcal{E}(x)$ along with $z_{m}, z_{s}$ is used to construct the forward diffusion process, producing the noisy latent $z_t$ at timestep $t$. The denoising network $\epsilon_\theta$, instantiated as our lightweight U-Net (see Fig.~\ref{fig:overall_pipeline}), is trained to predict the added noise $\epsilon$ given $z_t$, $t$.
To provide rich global semantic cues, we integrate the Latent Categories Guidance (LCG) paradigm \cite{xu2025pixelhacker}. LCG employs semantic embeddings to extract latent category distribution from unmasked images during training. These embeddings, denoted as $\mathbf{E}_\text{LCG} \in \mathbb{R}^{K \times D}$, act as external global priors injected into the denoising network via cross-attention mechanism. We set PixelHacker \cite{xu2025pixelhacker}, a SOTA LCG-based model, as our teacher network. Our goal is to forge a student model that retains the benefits of LDM and LCG while achieving extreme architectural efficiency.

\begin{table*}[t]
\centering
\caption{\textbf{Empirical Analysis of Representation Bottleneck and Architectural Synergy.} Proving the necessity of our knowledge distillation and the synergy of components. Models are evaluated on the Places2 (Test) benchmark using the 18K training checkpoint (evaluation details in Sec.~\ref{subsubsec:compare_details}). CA: standard cross-attention. L$\lambda$/I$\lambda$: Our Local/Interactive-$\lambda$ modules (details in Sec.~\ref{sec:lambda_modules}). KD: \ding{55} = standard prediction loss; \ding{51} = our knowledge distillation, which further verifies its validity in Tab.~\ref{tab: abla_distill}.
}
\label{tab:rebuttal_ablation}
\vspace{-0.5em}
\resizebox{0.73\linewidth}{!}{
\setlength{\tabcolsep}{6pt}
\begin{tabular}{c| >{\columncolor{TabGray}} c |c|cc| c c}
\toprule
\textbf{Exp} & \textbf{Arch} & \textbf{KD} & \textbf{FID} $\downarrow$ & \textbf{LPIPS} $\downarrow$ & \textbf{Param} $\downarrow$ & \textbf{GFLOPs} $\downarrow$ \\
\midrule
{\CircledNumber{1}} & GLA-CA-FFN, Conv & \ding{55} & 32.75 & 0.298 & \colorednumber{526}M & \colorednumberflops{314.30} \\
\midrule
{\CircledNumber{2}} & \textcolor{AblationOrangeWord}{L$\lambda$}-CA-FFN, Conv & \ding{55} & 37.65 & 0.325 & \colorednumber{496}M & \colorednumberflops{318.90} \\
{\CircledNumber{3}} & GLA-\textcolor{AblationOrangeWord}{I$\lambda$}-FFN, Conv & \ding{55} & 36.91 & 0.312 & \colorednumber{514}M & \colorednumberflops{307.91} \\
{\CircledNumber{4}} & GLA-CA-\textcolor{AblationOrangeWord}{MixFFN}, Conv & \ding{55} & 35.24 & 0.301 & \colorednumber{478}M & \colorednumberflops{286.51} \\
{\CircledNumber{5}} & GLA-CA-FFN, \textcolor{AblationOrangeWord}{DWConv} & \ding{55} & 43.58 & 0.341 & \colorednumber{315}M & \colorednumberflops{183.59} \\
\midrule
\midrule
{\CircledNumber{6}} & \textcolor{AblationOrangeWord}{L$\lambda$-I$\lambda$}-FFN, Conv & \ding{55} & 33.21 & 0.286 & \colorednumber{485}M & \colorednumberflops{312.50} \\
{\CircledNumber{7}} & \textcolor{AblationOrangeWord}{L$\lambda$-I$\lambda$}-FFN, Conv & \textcolor{AblationOrangeWord}{\ding{51}} & 24.73 & 0.257 & \colorednumber{485}M & \colorednumberflops{312.50} \\
\midrule
{\CircledNumber{8}} & \textcolor{AblationOrangeWord}{L$\lambda$-I$\lambda$}-FFN, \textcolor{AblationOrangeWord}{DWConv} & \textcolor{AblationOrangeWord}{\ding{51}} & 25.86 & 0.262 & \colorednumber{274}M & \colorednumberflops{181.79} \\
{\CircledNumber{\textbf{9}}} & \textbf{\textcolor{AblationOrangeWord}{L$\lambda$-I$\lambda$-MixFFN, DWConv}} & \textbf{\textcolor{AblationOrangeWord}{\ding{51}}} & \textbf{26.43} & \textbf{0.258} & \textbf{\colorednumber{226}M} & \textbf{\colorednumberflops{154.00}} \\
\midrule
{\CircledNumberX{10}} & \textcolor{AblationOrangeWord}{L$\lambda$-I$\lambda$-MixFFN, DWConv} & \ding{55} & 33.42 & 0.312 & \colorednumber{226}M & \colorednumberflops{154.00} \\
\midrule
\midrule
{\CircledNumberX{11}} & GLA-CA-FFN, Conv & \textcolor{AblationOrangeWord}{\ding{51}} & 26.81 & 0.262 & \colorednumber{526}M & \colorednumberflops{314.30} \\
{\CircledNumberX{12}} & \textcolor{AblationOrangeWord}{L$\lambda$}-CA-FFN, Conv & \textcolor{AblationOrangeWord}{\ding{51}} & 28.94 & 0.283 & \colorednumber{496}M & \colorednumberflops{318.90}  \\
{\CircledNumberX{13}} & GLA-\textcolor{AblationOrangeWord}{I$\lambda$}-FFN, Conv & \textcolor{AblationOrangeWord}{\ding{51}} & 26.46 & 0.265 & \colorednumber{514}M & \colorednumberflops{307.91} \\
{\CircledNumberX{14}} & GLA-CA-\textcolor{AblationOrangeWord}{MixFFN}, Conv & \textcolor{AblationOrangeWord}{\ding{51}} & 27.35 & 0.251 & \colorednumber{478}M & \colorednumberflops{286.51} \\
{\CircledNumberX{15}} & GLA-CA-FFN, \textcolor{AblationOrangeWord}{DWConv} & \textcolor{AblationOrangeWord}{\ding{51}} & 29.41 & 0.285 & \colorednumber{315}M & \colorednumberflops{183.59} \\
\bottomrule
\end{tabular}
}
\vspace{-2.0em}
\end{table*}

\begin{figure}[tb]
\centering
\begin{minipage}[p]{0.48\linewidth}
\includegraphics[width=1.0\linewidth]{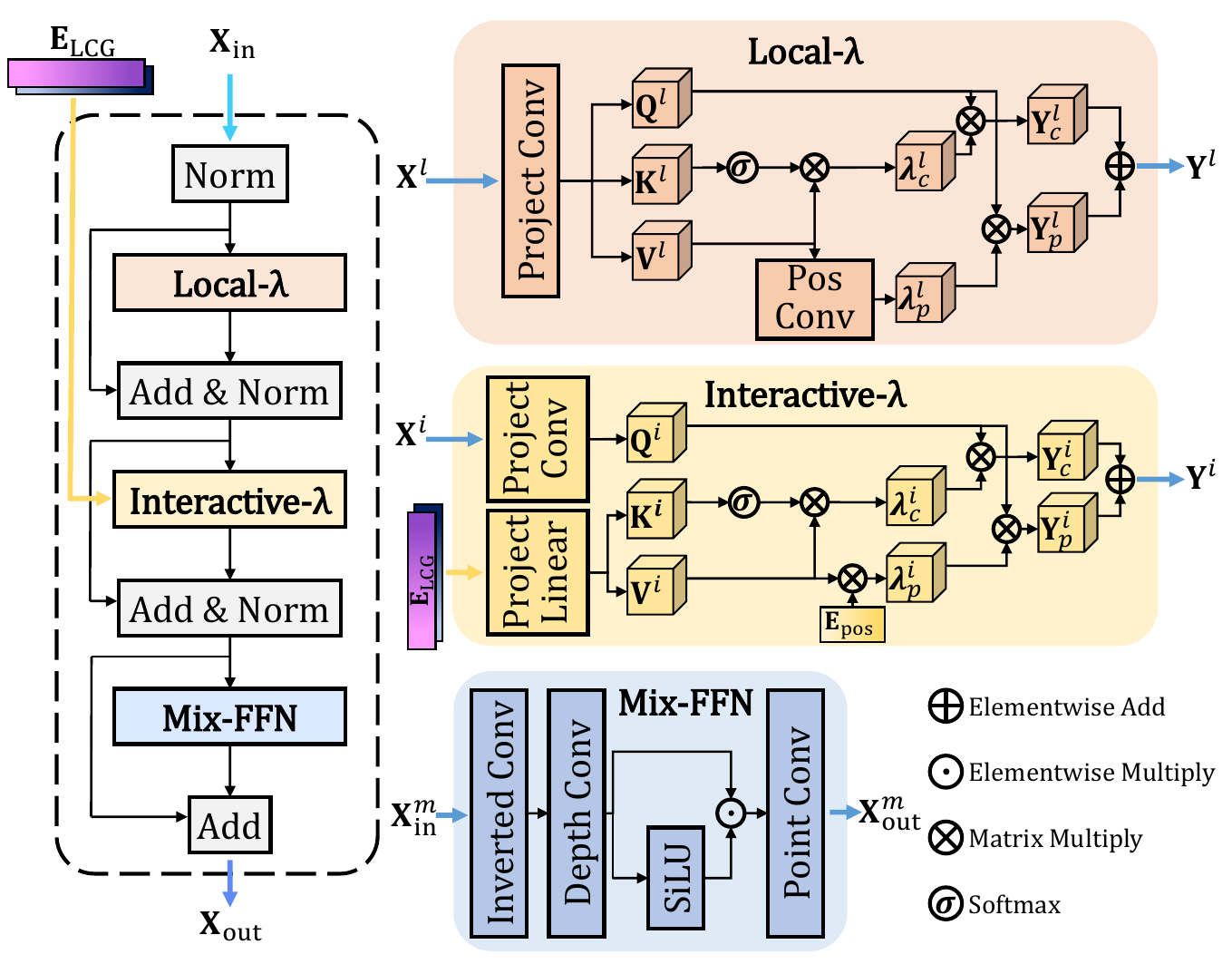}
\caption{
\textbf{Detailed architecture of the Local $\boldsymbol{\lambda}$ Mix Interaction ($\boldsymbol{L\lambda MI}$) Block.}
The left panel illustrates the overall architecture, comprising three main submodules: Local-$\lambda$, Interactive-$\lambda$, and Mix-FFN. We elaborate on their mathematical formulations in Sec.~\ref{sec:lambda_modules}.
}
\label{fig:local_lambda_mix_interaction}
\end{minipage}
\hfill
\begin{minipage}[p]{0.48\linewidth}
\includegraphics[width=1.0\linewidth]{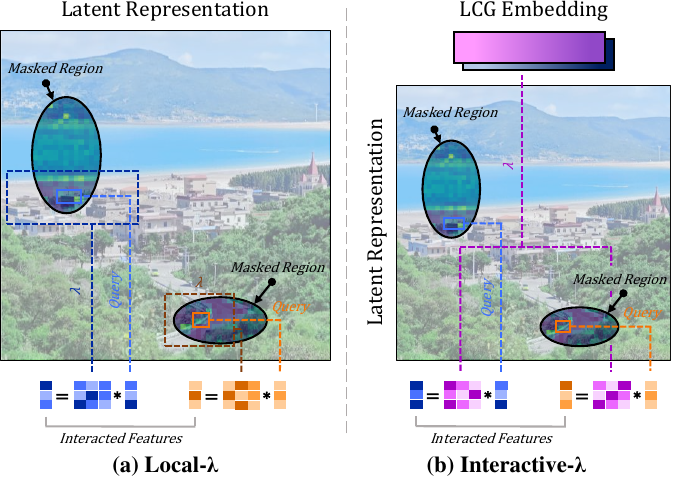}
\caption{
\textbf{Illustration of local context aggregation (Local-$\lambda$) and cross-embedding interaction (Interactive-$\lambda$) in the latent domain.}
In both modules, $\lambda$ efficiently summarizes either spatial contexts or the global prior $\mathbf{E}_\text{LCG}$ into a fixed-size linear matrix, bypassing memory-intensive attention calculations.
}
\label{fig:local_global_lambda}
\end{minipage}
\vspace{-0.8em}
\end{figure}

\vspace{-0.5em}
\subsection{Architecture Evolution: The Path to Moebius}\label{sec:architecture_evolution}
\vspace{-0.5em}
To achieve extreme architectural efficiency, we systematically restructure the diffusion backbone. In this subsection, we first analyze the representation bottleneck caused by naive structural compression, and subsequently detail our formulaically grounded solutions, as illustrated in Fig.~\ref{fig:local_lambda_mix_interaction} and Fig.~\ref{fig:local_global_lambda}.
\vspace{-0.7em}

\vspace{-1.5em}
\subsubsection{Empirical Motivation: The Representation Bottleneck.}\label{sec:motivation}
A natural starting point for compression is to simplify the macro-architecture of the teacher model. Our empirical analysis reveals that removing the redundant final downsampling stage of PixelHacker provides a solid baseline. This modification significantly reduces the parameter count from 862M to 526M while only marginally sacrificing generation quality (FID: 32.17 $\rightarrow$ 32.75, LPIPS: 0.292 $\rightarrow$ 0.298), yielding the solid baseline corresponding to Exp \textcircled{\small 1} in Tab.~\ref{tab:rebuttal_ablation}. However, pushing compression further at the micro-architecture level poses severe challenges. A straightforward approach would be substituting standard convolutions, attention mechanisms (namely, the GLA \cite{yang2024gla} used for self-attention, and the standard cross-attention), and FFN with their efficiency variants \cite{Sandler2018MobileNetV2, xie2024sana, xie2021segformer}.

Our empirical observation reveals that such naive substitutions inevitably trigger a severe representation bottleneck. As shown in Tab.~\ref{tab:rebuttal_ablation} (Exp \textcircled{\small 2}-\textcircled{\small 5}), simply dropping these lightweight operators directly into the backbone without architectural synergy leads to catastrophic degradation in generation quality (e.g., FID deteriorates from 32.75 to over 43.58). This bottleneck arises from an intrinsic flaw: lightweight operators inherently suffer from constrained representational capacity, struggling to model the rigorous semantic reasoning required for inpainting. In addition, GLA \cite{yang2024gla} in the baseline is highly efficient for self-attention but lacks a mathematical formulation to perform cross-attention. This intrinsic limitation completely obstructs the synergistic architectural optimization of self- and cross-attention operations.

\vspace{-1.6em}
\subsubsection{Local and Interactive $\lambda$ Modules.}\label{sec:lambda_modules}
To break through the representation bottleneck and efficiently integrate $\mathbf{E}_\text{LCG}$, we introduce the Local-$\lambda$ and Interactive-$\lambda$ modules. Our core insight is to bypass memory-intensive dot-product attention maps by summarizing contextual and semantic information into fixed-size linear matrices (denoted as $\lambda$), allowing for linear-complexity interactions.

\vspace{-0.5em}
\paragraph{\underline{Local $\lambda$ for Self-Attention Equivalent.}}
The Local-$\lambda$ module aggregates intra-image semantic and spatial contexts. Given an input latent feature $\mathbf{X}^l$ shaped as $B\times H^\prime\times W^\prime\times C$, we first project it into multi-query $\mathbf{Q}^l$, key $\mathbf{K}^l$, and value $\mathbf{V}^l$ via $1\times1$ convolutions and batch normalizations. Instead of computing quadratic attention maps, we construct a semantic content mapping $\boldsymbol{\lambda}^l_c$ and a positional mapping $\boldsymbol{\lambda}^l_p$:
\begin{equation}
\begin{gathered}
\boldsymbol{\lambda}^l_c = \mathrm{softmax}(\mathbf{K}^l)^\top \mathbf{V}^l, \quad
\boldsymbol{\lambda}^l_p = \mathrm{Conv3D}^{\text{pos}}_{1\times r\times r}(\mathbf{V}^l), 
\end{gathered}
\end{equation}
where $r$ defines the local perception window size. The query $\mathbf{Q}^l$ then linearly interacts with these two compact matrices to produce the final self-aggregated output $\mathbf{Y}^l = \mathbf{Q}^l \boldsymbol{\lambda}^l_c + \mathbf{Q}^l \boldsymbol{\lambda}^l_p$. This dual aggregation elegantly integrates local spatial continuousness and semantic content while maintaining linear complexity.

\vspace{-0.5em}
\paragraph{\underline{Interactive $\lambda$ for Cross-Attention Equivalent.}}
To address the critical inability of GLA in handling cross-attention, we propose the Interactive-$\lambda$ module. This module is specifically formulated to interact the latent representations with the global semantic prior $\mathbf{E}_\text{LCG}$. We project the latent representation $\mathbf{X}^i$ into query $\mathbf{Q}^i$, while projecting $\mathbf{E}_\text{LCG}$ into key $\mathbf{K}^i$ and value $\mathbf{V}^i$. 
Since $\mathbf{E}_\text{LCG}$ possesses a much smaller spatial scale than the latent representation, establishing a precise spatial-semantic correspondence is challenging. To resolve this, we introduce a lightweight positional embedding $\mathbf{E}_\text{pos}$ to inject explicit spatial layout information into the values, yielding the positional mapping $\boldsymbol{\lambda}^i_p$. The final interactive output $\mathbf{Y}^i$ is aggregated as:
\begin{equation}
\begin{gathered}
 \boldsymbol{\lambda}^i_c =  \mathrm{softmax}(\mathbf{K}^i)^\top \mathbf{V}^i, \quad
 \boldsymbol{\lambda}^i_p = \mathbf{E}_\text{pos} \mathbf{V}^i, \quad
 \mathbf{Y}^i = \mathbf{Q}^i \boldsymbol{\lambda}^i_c + \mathbf{Q}^i \boldsymbol{\lambda}^i_p.
\end{gathered}
\end{equation}

\vspace{-0.5em}
By framing cross-attention through the lens of fixed-size matrices, the Interac- tive-$\lambda$ module successfully integrates external semantic priors at a fraction of the traditional computational cost, solving the architectural impasse of GLA. Crucially, the joint integration of Local-$\lambda$ and Interactive-$\lambda$ establishes a highly efficient interaction paradigm. As validated in Tab.~\ref{tab:rebuttal_ablation} (Exp \textcircled{\small 1} $\rightarrow$ \textcircled{\small 6}), completely replacing the baseline's attention mechanisms with our dual $\lambda$ modules reduces the parameter count (526M $\rightarrow$ 485M) and computational overhead, while maintaining highly competitive generation quality (FID: 32.75 $\rightarrow$ 33.21) and even improving perceptual alignment (LPIPS: 0.298 $\rightarrow$ 0.286).

\vspace{-1em}
\subsubsection{The $L\lambda MI$ Block and Lightweight Instantiation.}\label{sec:llm_instantiation}
In a standard latent diffusion U-Net, the backbone consists of two primary components: convolutional residual blocks for spatial feature extraction, and attention blocks for token interaction. To align with the extreme efficiency of our $\lambda$ modules and achieve a holistic lightweight instantiation, we comprehensively restructure both.

\vspace{-0.5em}
\paragraph{\underline{Pushing Extreme Compression via DWConv and Mix-FFN.}}
For spatial feature extraction, we replace the computationally heavy standard convolutional blocks with Depthwise Residual Blocks (DW.Res) \cite{Sandler2018MobileNetV2}. While this substitution achieves massive parameter savings (Tab.~\ref{tab:rebuttal_ablation}, Exp \textcircled{\small 7} $\rightarrow$ \textcircled{\small 8}), the standard FFNs within the interaction blocks still dominate the parameter budget. To push Moebius into the extreme lightweight regime (ie., $<2\%$ of the 11.9B parameters of the industrial SOTA, FLUX.1-Fill-Dev), compressing the FFNs is unavoidable. Therefore, we integrate Mix-FFN \cite{xie2024sana, xie2025sana}, which replaces dense linear projections with a highly efficient depthwise-augmented structure. As empirically shown (Exp \textcircled{\small 8} $\rightarrow$ \textcircled{\small \textbf{9}}), integrating Mix-FFN slashes an additional 48M parameters and 27 GFLOPs. Although this aggressive compression introduces a slight trade-off in FID (25.86 $\rightarrow$ 26.43), it impressively preserves the strict perceptual quality (LPIPS: 0.262 $\rightarrow$ 0.258). This demonstrates that Mix-FFN is an excellent choice for crossing the extreme efficiency threshold without collapsing the representation.

\vspace{-0.5em}
\paragraph{\underline{Formulation of the $L\lambda MI$ Block.}}
By elegantly cascading the proposed Local-$\lambda$, Interactive-$\lambda$, and Mix-FFN, we formulate our ultimate building block: the Local-$\lambda$ Mix Interaction ($L\lambda MI$) block. Given an input latent feature $\mathbf{X}_{in}$, the forward pass of the $L\lambda MI$ block is mathematically defined as:
\begin{equation}
\begin{aligned}
\mathbf{X}_1 &= \text{Local-}\lambda(\text{LN}(\mathbf{X}_{in})) + \mathbf{X}_{in}, \\
\mathbf{X}_2 &= \text{Interactive-}\lambda(\text{LN}(\mathbf{X}_1), \mathbf{E}_\text{LCG}) + \mathbf{X}_1, \\
\mathbf{X}_{out} &= \text{Mix-FFN}(\text{LN}(\mathbf{X}_2)) + \mathbf{X}_2,
\end{aligned}
\end{equation}
where $\text{LN}(\cdot)$ denotes Layer Normalization. This block successfully replaces the cumbersome spatial transformer blocks found in heavy diffusion models. 

\vspace{-0.5em}
\paragraph{\underline{Architectural Synergy and the Capacity Dilemma.}}\label{para:Architectural_Synergy_and_Capacity_Dilemma}
We construct the Moebius denoising U-Net by strategically stacking DWConv blocks and $L\lambda MI$ blocks across varying resolutions. As empirically validated in Tab.~\ref{tab:rebuttal_ablation} (Exp \textcircled{\small \textbf{9}}), this specific architectural combination achieves an optimal structural synergy. It drastically compresses the parameters to 0.22B and FLOPs to 0.154T, while maintaining a competitive generation quality under our fully equipped optimization scheme. 
\indent However, structural efficiency comes at an inherent cost. As observed in Exp \textcircled{\scriptsize 10} (Tab.~\ref{tab:rebuttal_ablation}), operating at this extreme compression scale solely with standard prediction loss bounds the absolute representational ceiling of the model (yielding a degraded FID of 33.42). To fully unlock the potential of the $L\lambda MI$ architecture, recover the lost capacity, and bridge the massive performance gap to 10B-level models, an advanced optimization strategy is imperative. This motivates our proposed multi-granularity distillation.

\begin{wrapfigure}{R}{0.5\textwidth}
\vspace{-2.2em}
\centering
\includegraphics[width=0.98\linewidth]{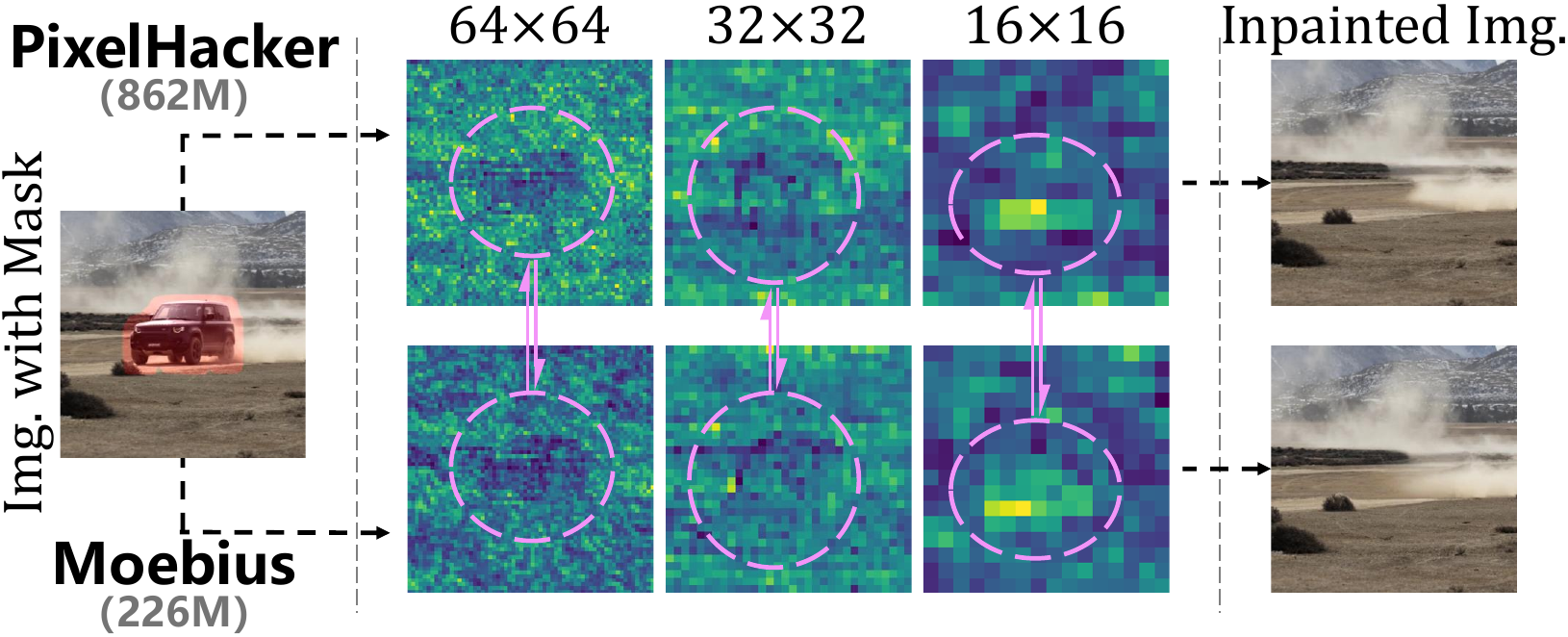}
\caption{
\textbf{Small feature spaces can still maintain high representation quality.}
Moebius (0.22B) exhibits highly similar activation maps to the teacher model, PixelHacker (0.86B), across multiple spatial granularities, demonstrating that it maintains consistent representational quality despite a severely compressed (4$\times$ smaller) architecture. This validates the optimal synergy between our lightweight design and the adaptive multi-granularity distillation.
}
\vspace{-1em}
\label{fig:representation_align}
\end{wrapfigure}

\vspace{-1em}
\subsection{Adaptive Multi-Granularity Distillation}\label{sec:distillation}
\vspace{-0.5em}

As discussed in Sec.~\ref{para:Architectural_Synergy_and_Capacity_Dilemma}, while the $L\lambda MI$ architecture successfully pushes the model into the extreme lightweight regime, this profound structural compression inevitably restricts the absolute representational capacity of the network. To bridge this capacity gap and achieve optimal architectural synergy, we introduce a multi-granularity distillation strategy. 
Crucially, prior works \cite{kang2024diffusion2gan, zhang2018lpips, reda2022film} have shown that perceptual constraints can significantly preserve important structural details. However, the memory overhead of decoding high-resolution latents back into the pixel space during training is computationally prohibitive for a lightweight framework. To maintain extreme training efficiency, we perform the entire distillation—including perceptual alignment—strictly within the latent space. As demonstrated in Fig.~\ref{fig:representation_align}, Moebius, empowered by this distillation strategy, successfully preserves high-quality representations across multiple granularities.

\vspace{-0.5em}
\vspace{-0.9em}
\subsubsection{Multi-Granularity Distillation Objectives.}
We adopt a standard Teacher-Student paradigm. The teacher model $T$ is the officially pretrained PixelHacker \cite{xu2025pixelhacker}, a high-capacity diffusion backbone equipped with LCG guidance, serving as the upper-bound performance reference. The student model $S$ is our lightweight Moebius. To transfer the semantic and generative capabilities from $T$ to $S$, we define the following optimization objectives:

\vspace{-0.5em}
\vspace{-0.9em}
\paragraph{\underline{Coarse-Grained Distillation.}}
At the coarse spatial granularity ($16\times16$), we enforce an element-wise alignment within the intermediate bottleneck. We extract the teacher's latent representation $\hat{x}_\mathrm{C\_T}$ after its first upsampling block, and align it with the student's latent representation $\hat{x}_\mathrm{C\_S}$ after its last downsampling block, formulating the coarse-grained distillation loss: $\mathcal{L}_\mathrm{C\_KD} = \| \hat{x}_\mathrm{C\_T} - \hat{x}_\mathrm{C\_S} \|^2_2$.

\vspace{-0.5em}
\vspace{-0.5em}
\paragraph{\underline{Fine-Grained Distillation and Task Supervision.}}
At the fine spatial granularity ($64\times64$), we supervise the final latent output. We first define a standard task loss between the student's prediction $\hat{x}_\mathrm{S}$ and the ground-truth latent noise $x_0$: $\mathcal{L}_\mathrm{task} =  \| x_0 - \hat{x}_\mathrm{S} \|^2_2$.
Simultaneously, we align the final predictions of the teacher $\hat{x}_\mathrm{T}$ and the student via an L2 distillation loss: $\mathcal{L}_\mathrm{F\_KD} = \| \hat{x}_\mathrm{T} - \hat{x}_\mathrm{S} \|^2_2$.

\vspace{-0.5em}
\vspace{-0.25em}
\paragraph{\underline{Latent Perceptual Distillation.}} 
To enhance the perceptual quality of the generated outputs without incurring the expensive pixel-space decoding costs, we employ the memory-efficient E-LatentLPIPS \cite{kang2024diffusion2gan} as a perceptual constraint directly in the latent space: $\mathcal{L}_\mathrm{perceptual} =  d_\mathrm{E\_LatentLPIPS}(x_0, \hat{x}_\mathrm{S})$.
This loss aligns perceptual features entirely within the latent domain, significantly reducing memory usage and perfectly complementing our lightweight training philosophy.

\vspace{-0.5em}
\subsubsection{Adaptive Gradient-Based Balance.}
In practice, optimizing the student with the aforementioned heterogeneous objectives introduces a severe convergence challenge. We observe that the losses at coarse and fine granularities differ greatly in magnitude and gradient contribution. Relying on static hyperparameter weighting makes it extremely difficult to balance convergence and generation quality.
Inspired by GAN Loss \cite{esser2021VQGAN}, we propose an adaptive mechanism that dynamically adjusts the loss weights according to their gradient norms with respect to key parameter sets. Let $G(\mathcal{L},\theta)$ denote the gradient norm (e.g., L2 norm) of loss $\mathcal{L}$ at parameter set $\theta$.
For intra-layer balance at the fine granularity, we define the adaptive weights $\mathcal{W}_{\text{F\_KD}}$ and $\mathcal{W}_{\text{perceptual}}$ based on parameters of the final output layer, $\theta_\text{F}$. The weighted fine-grained output loss $\mathcal{L}_{\text{out}}$ is formulated~as:
\begin{equation}
\begin{gathered}
\mathcal{W}_{\text{F\_KD}} =  \frac{||G(\mathcal{L}_{\text{task}}, \theta_\text{F})||^2_2}{||G(\mathcal{L}_{\text{F\_KD}}, \theta_\text{F})||^2_2}, \quad
\mathcal{W}_{\text{perceptual}} =  \frac{||G(\mathcal{L}_{\text{task}}, \theta_\text{F})||^2_2}{||G(\mathcal{L}_{\text{perceptual}}, \theta_\text{F})||^2_2}, \\
\mathcal{L}_{\text{out}} = \mathcal{L}_{\text{task}} + \mathcal{W}_{\text{F\_KD}} \cdot \mathcal{L}_{\text{F\_KD}}  + \mathcal{W}_{\text{perceptual}} \cdot \mathcal{L}_{\text{perceptual}}.
\end{gathered}
\end{equation}

To balance the contributions across different granularities, we compute a cross-granularity weight $\mathcal{W}_{\text{C\_task}}$. This is based on the gradient norms with respect to the intermediate feature parameters $\theta_\text{C}$ (i.e., the last layer before $\hat{x}_{\text{C\_S}}$). The total training objective $\mathcal{L}_{\text{total}}$ is defined as: 
\begin{equation}
\begin{gathered}
    \mathcal{W}_{\text{C\_task}} = \frac{||G(\mathcal{L}_{\text{C\_KD}}, \theta_\text{C})||^2_2}{||G(\mathcal{L}_{\text{out}}, \theta_\text{C})||^2_2}, \quad
    \mathcal{L}_{\text{total}} = \mathcal{L}_{\text{C\_KD}} + \mathcal{W}_{\text{C\_task}} \cdot \mathcal{L}_{\text{out}}.
\end{gathered}
\end{equation}

This adaptive balancing mechanism effectively stabilizes the joint optimization and alleviates the need for extensive manual hyperparameter tuning, allowing the lightweight student to converge rapidly and smoothly.

\vspace{-0.5em}
\section{Experiments}\label{sec:experiments}
\vspace{-0.5em}

\subsection{Experimental Setup}\label{subsec:setup}

\vspace{-0.4em}
\subsubsection{Implementation Details.}\label{subsubsec:impl}
We adopt the officially pretrained PixelHacker \cite{xu2025pixelhacker} as our teacher model, sharing the same LCG embedding size, 512$\times$512 input resolution, and SDXL VAE encoder \cite{podell2023sdxl}. For the Local-$\lambda$ module, the perception window size $r$ is set to 15. We employ the Muon optimizer \cite{jordan2024muon, kimiteam2025kimik2} with a weight decay of 0.1. 
The multi-granularity distillation is conducted on 16 NVIDIA L40S GPUs with a total batch size of 768 for 138K iterations in BF16 precision. The learning rate initiates at 2e-4 and decays by a factor of 0.1 at 111K and 129K iterations. Subsequently, Moebius is fine-tuned on individual benchmarks using NVIDIA RTX 3090 GPUs. Following common practices \cite{suvorov2021lama, Sargsyan2023migan, li2022mat}, we fine-tune on Places2 \cite{zhou2017places} (1.8M images, 4 GPUs, batch size 88, 51K iters), CelebA-HQ \cite{karras2018celebahq} (24K images, 2 GPUs, batch size 44, 60K iters), and FFHQ \cite{karras2018stylegan_ffhq} (60K images, 4 GPUs, batch size 88, 117K iters). 

\vspace{-1em}
\vspace{-0.4em}
\subsubsection{Evaluation Protocols.}\label{subsubsec:compare_details}
We rigorously evaluate Moebius across natural \cite{zhou2017places} and portrait \cite{karras2018stylegan_ffhq, karras2018celebahq} domains, reporting FID \cite{huesel2017FID} and LPIPS \cite{zhang2018lpips} following common practice.
\textbf{Places2 (Test)}: following PowerPaint \cite{zhuang2023powerpaint}, 10K test subset, 512$\times$512, 40–50\% masks;
\textbf{Places2 (Large/Small)}: following MAT \cite{li2022mat}, 36.5K validation images, 512$\times$512, large/small masks;
\textbf{Places2 (256)}: following MI-GAN \cite{Sargsyan2023migan}, 36.5K validation images, 256$\times$256, free-form masks;
\textbf{CelebA-HQ (512)}: following MAT \cite{li2022mat}, 3k images, 512$\times$512, large masks;
\textbf{FFHQ (256)}: following MI-GAN \cite{Sargsyan2023migan}, 10K images, 256$\times$256, LaMa-style masks \cite{suvorov2021lama}).

\vspace{-1em}
\vspace{-0.4em}
\subsubsection{Baselines \& Fair Efficiency Profiling.} 
We conduct extensive comparisons against SOTA academic task-specific specialists—encompassing top-ranked methods from the Papers-With-Code platform\cite{paperwithcode} alongside the latest advancements \cite{xu2025pixelhacker, Li2025RoRem, Chen2024LCI, manukyan2023hdpainter}—as well as massive industrial zero-shot generalists \cite{flux2024, esser2024SD3}. In all tables, "$\dag$" denotes results reported in the original papers \cite{xu2025pixelhacker, li2022mat, Sargsyan2023migan, zhuang2023powerpaint}, while others are reproduced using official weights and code. To ensure fairness in efficiency profiling, all single-step inference latencies are measured under a strictly standardized environment: a single L40S GPU, batch size 1 at 512$\times$512.

\vspace{-0.5em}
\subsection{Main Results: Bridging the Scale Gap}\label{subsec:main_results}

\vspace{-0.5em}
\paragraph{\underline{Extreme Architectural Efficiency.}}
As detailed in Tab.~\ref{tab:total_natural}, Moebius establishes a new efficiency standard. Operating with a mere 0.226B parameters, it achieves an outstanding inference speed of 26.01 ms/step, significantly outperforming all diffusion-based competitors. When contrasted with 10B-level industrial generalists like FLUX.1-Fill-Dev (11.9B) and SD3.5 Large-Inp. (8.05B), Moebius operates with less than \textbf{2\%} to \textbf{3\%} of their parameter budget and delivers a $\boldsymbol{6\times}$ acceleration in single-step latency. When factoring in total sampling steps, this efficiency gap widens substantially (Tab.~\ref{tab:cover}), highlighting the massive redundancy inherent in industrial foundational models for specific restoration tasks.

\begin{table}[tb]
\caption{
\textbf{Quantitative comparison with SOTA academic and industrial methods on Places2 (Test/Large/Small/256, natural scene).} 
“$\dag$”: results reported in original papers \cite{xu2025pixelhacker, li2022mat, Sargsyan2023migan, zhuang2023powerpaint}.
“Indu.” denotes industrial.
Benchmark details in Sec.~\ref{subsubsec:compare_details}.
}
\vspace{-0.5em}
\centering
\resizebox{1.0\linewidth}{!}{
\begin{tabular}{ >{\hspace{2pt}}c<{\hspace{2pt}}|l|cc|cc|cc|cc|cc}
\toprule
\multicolumn{2}{c|}{\multirow{2}{*}{\textbf{Method}}} 
& \multicolumn{2}{c|}{\textbf{Places2 (Test)}}  
& \multicolumn{2}{c|}{\textbf{Places2 (Large)}} 
& \multicolumn{2}{c|}{\textbf{Places2 (Small)}}
& \multicolumn{2}{c|}{\textbf{Places2 (256)}} 
& \textbf{Param.} & \textbf{Latency}  \\
\multicolumn{2}{c|}{}
&       $\mathbf{FID}\downarrow$    &     $\mathbf{LPIPS}\downarrow$  
&       $\mathbf{FID}\downarrow$    &     $\mathbf{LPIPS}\downarrow$  
&       $\mathbf{FID}\downarrow$    &     $\mathbf{LPIPS}\downarrow$  
&       $\mathbf{FID}\downarrow$    &     $\mathbf{LPIPS}\downarrow$ 
& $\mathbf{(\times 10^9)}\downarrow$ & $\mathbf{(ms/step)}\downarrow$ \\
\midrule
\rowcolor{TabGray} \multicolumn{12}{c}{ \textit{Non-Diffusion-based Methods} }  \\
\midrule
\multirow{9}{*}{\rotatebox{90}{\scriptsize\textbf{Academic}}} 
&LaMa \cite{suvorov2021lama}      & 21.07\dg\more{122} & 0.213\dg\more{3} &-&-&-&-& 22.00\dg\more{56} & 0.378\dg\less{20} &-&- \\
&MI-GAN \cite{Sargsyan2023migan}  & 14.36\dg\more{51} & 0.239\dg\more{15} &-&-&-&-& 11.83\dg\less{16} & 0.394\dg\less{16} &-&- \\
&MADF \cite{zhu2021MADF}          & - & - &  7.53\dg\more{220}  & 0.181\dg\more{5} & 2.24\dg\more{143} & 0.095\dg\more{4} &-&-&-&- \\
&AOT-GAN \cite{zeng2023AOTGAN}    & - & - & 10.64\dg\more{353}  & 0.195\dg\more{13} & 3.19\dg\more{247} & 0.101\dg\more{11} &-&-&-&- \\
&HiFill \cite{yi2020hifill}       & - & - & 28.92\dg\more{1131} & 0.284\dg\more{64} & 7.94\dg\more{763} & 0.148\dg\more{63} & 81.27\dg\more{475} & 0.488\dg\more{4} &-&- \\
&DeepFillv2 \cite{yu2019deepfillv2} & - & - & 9.27\dg\more{294} & 0.213\dg\more{23} & 3.02\dg\more{228} & 0.113\dg\more{24} &-&-&-&- \\
&EdgeConnect \cite{naz2019EC}       & - & - &12.66\dg\more{439} & 0.275\dg\more{59} & 4.03\dg\more{338} & 0.114\dg\more{25} &-&-&-&- \\
&MAT \cite{li2022mat}               & 9.27\dg\less{2} & 0.211\dg\more{2} & 2.90\dg\more{23} & 0.189\dg\more{9} & 1.07\dg\more{16} & 0.099\dg\more{9} & 14.38\dg\more{2} & 0.394\dg\less{16} &-&-\\
&Latent-C.I. \cite{Chen2024LCI}     & 23.14\more{144} & 0.288\more{39} & 5.08\more{116} & 0.210\more{21} & 1.59\more{73} & 0.108\more{19} & 30.72\more{117} & 0.478\more{2} & 1.686\more{646} & - \\
\midrule
\rowcolor{TabGray} \multicolumn{12}{c}{ \textit{Diffusion-based Methods} } \\
\midrule
\multirow{7}{*}{\rotatebox{90}{\scriptsize\textbf{Academic}}}
& LDM  \cite{Rombach2022LDM}               & 21.42\dg\more{126}& 0.232\dg\more{12} & - & - & - & - &13.40\dg\less{5} & 0.385\dg\less{18} & 0.387\dg\more{71} & - \\
& SD \cite{Rombach2022LDM}                 & 19.73\dg\more{108}& 0.232\dg\more{12} &-&-&-&-&-&-& 0.860\more{281} & 57.07\more{119} \\
& PowerPaint \cite{zhuang2023powerpaint}   & 17.91\dg\more{89}& 0.223\dg\more{8}  &-&-&-&-&-&-& 0.860\more{281} & 56.58\more{118} \\
& HD-Painter \cite{manukyan2023hdpainter}  & 23.53\more{148}  & 0.322\more{56} & 8.07\more{243} & 0.293\more{69} & 4.49\more{388} & 0.203\more{123} & 37.10\more{163} & 0.527\more{12} & 0.860\more{281} & 58.61\more{125} \\
& DDNM  \cite{wang2022ddnm}                & 29.62\more{212} & 0.727\more{251} & 9.18\more{291} & 0.846\more{389} & 5.43\more{490} & 0.839\more{822} & 39.13\more{177} & 0.810\more{72} & 0.553\more{145} & 195.50\more{652} \\
& RoRem \cite{Li2025RoRem}                 & 24.17\more{155} & 0.297\more{43} & 5.88\more{150} & 0.251\more{45} & 1.78\more{93} & 0.133\more{46} & 28.63\more{103} & 0.468\less{0} & 2.567\more{1036} & 113.66\more{337} \\
& PixelHacker \cite{xu2025pixelhacker}      &  8.59\dg\less{9} & 0.203\dg\less{2} & 2.05\dg\less{13} & 0.169\dg\less{2} & 0.82\dg\less{11} & 0.088\dg\less{3}  & 9.25\dg\less{35} & 0.367\dg\less{22} & 0.862\more{281} & 46.89\more{80}  \\
\midrule
\multirow{2}{*}{\rotatebox{90}{\scriptsize\textbf{Indu.}}}
& SD3.5 Large-Inp. \cite{esser2024SD3}      & 37.33\more{294} & 0.237\more{14} & 10.94\more{366} & 0.202\more{17} & 3.02\more{228} & 0.105\more{15}  & 74.29\more{426} & 0.595\more{27} &  8.057\more{3465} & 151.02\more{481} \\
& FLUX.1-Fill-Dev \cite{flux2024}           &  8.02\less{15} & 0.279\more{35} & 1.86\less{21} & 0.179\more{3} & 0.94\more{2} & 0.099\more{9}  & 10.44\less{26} & 0.391\less{17} &  11.902\more{5166} & 161.01\more{519} \\
\midrule
\midrule
\rowcolor{AblationGreen} & {Moebius}     & 9.48 & 0.207 & 2.35 & 0.173 & 0.92 & 0.091 & 14.13 & 0.470 & 0.226 & 26.01 \\
\bottomrule
\end{tabular}
}
\label{tab:total_natural}
\vspace{-1.5em}
\end{table}

\begin{figure}[t]
\centering
\includegraphics[width=1.0\linewidth]{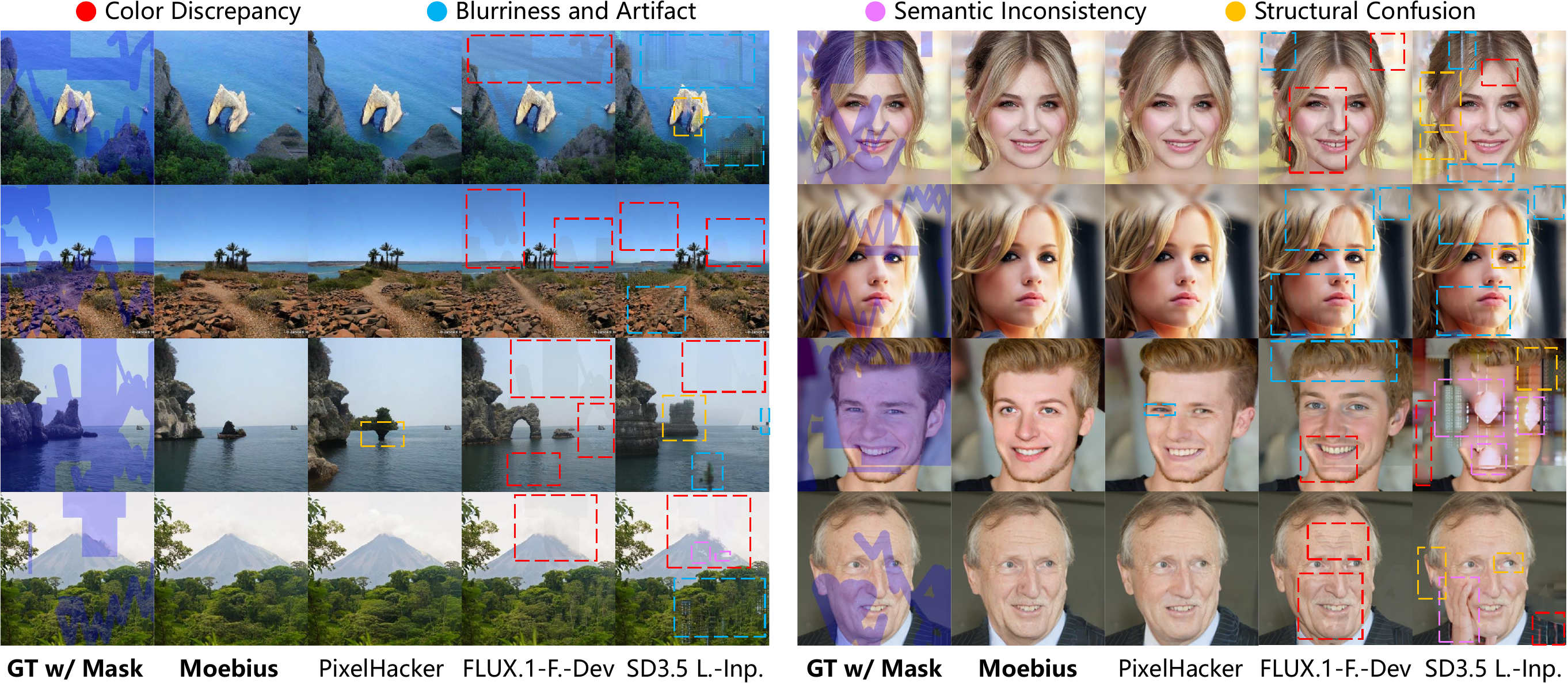}
\vspace{-1.8em}
\caption{
\textbf{Qualitative comparison with SOTA academic and industrial methods on natural and portrait scenes.} Left: Places2. Right, rows 1–2: CelebA-HQ; rows 3–4: FFHQ. 
Moebius delivers consistent contextual generation across natural and portrait domains, avoiding the common failure cases of other methods, including color discrepancies, blur, artifacts, semantic inconsistencies, and structural confusion.
}
\label{fig:qualitative_comparison}
\end{figure}

\vspace{-0.5em}
\paragraph{\underline{Performance on Natural Scenes.}}
Despite its extreme compactness, Moebius successfully bridges the capacity gap. On the natural-scene benchmarks (Tab.~\ref{tab:total_natural}), Moebius demonstrates highly competitive generation capabilities. When excluding its teacher model (PixelHacker), Moebius performs on par with the 10B-level industrial SOTA, FLUX.1-Fill-Dev, across various mask conditions, and even surpasses it on the Places2 (Small) benchmark with leading scores of 0.92 FID and 0.091 LPIPS. Furthermore, it significantly outperforms SD3.5 Large-Inpainting and the vast majority of academic methods. As qualitatively shown in Fig.~\ref{fig:qualitative_comparison} (Left), industrial models frequently suffer from color discrepancies and structural artifacts in fine-grained textures (e.g., foliage, water). In contrast, Moebius leverages its task-specific refinement and global semantic priors to deliver highly coherent restorations, matching the fidelity of its heavy teacher.

\begin{wraptable}{R}{0.5\textwidth}
\vspace{-3.5em}
\caption{
\textbf{Quantitative comparison with SOTA academic and industrial methods on CelebA-HQ and FFHQ (portrait scene).}
“$\dag$”: reported in original papers \cite{xu2025pixelhacker, li2022mat, Sargsyan2023migan, zhuang2023powerpaint}.
“Indu.” denotes industrial.
Benchmark details in Sec.~\ref{subsubsec:compare_details}.
}
\vspace{1em}
\centering
\resizebox{1.0\linewidth}{!}{
\begin{tabular}{>{\hspace{2pt}}c<{\hspace{2pt}}|l|cc|cc}
\toprule
\multicolumn{2}{c|}{\multirow{2}{*}{\textbf{Method}}}
& \multicolumn{2}{c|}{\textbf{CelebA-HQ (512)}} 
& \multicolumn{2}{c }{\textbf{FFHQ (256)}} \\
\multicolumn{2}{c|}{}
&  $\mathbf{FID}\downarrow$ & $\mathbf{LPIPS}\downarrow$  
&  $\mathbf{FID}\downarrow$ & $\mathbf{LPIPS}\downarrow$ \\
\midrule
\rowcolor{TabGray} \multicolumn{6}{c}{ \textit{Non-Diffusion-based Methods} }  \\
\midrule
\multirow{10}{*}{\rotatebox{90}{\scriptsize\textbf{Academic}}}
&CoModGAN \cite{zhao2021comodgan}   &  5.65\dg\more{5}   & 0.140\dg\more{15} &    -               &   -   \\
&LaMa \cite{suvorov2021lama}        &  8.15\dg\more{51}  & 0.143\dg\more{17} & 32.45\dg\more{298} & 0.294\dg\more{27} \\
&ICT  \cite{wan2021ICT}             & 12.84\dg\more{138} & 0.195\dg\more{60} &    -               &   -   \\
&MADF \cite{zhu2021MADF}            &  6.83\dg\more{27}  & 0.130\dg\more{7}  &    -               &   -   \\
&AOT-GAN \cite{zeng2023AOTGAN}      & 10.82\dg\more{101} & 0.145\dg\more{19} &    -               &   -   \\
&DeepFillv2 \cite{yu2019deepfillv2} & 24.42\dg\more{353} & 0.221\dg\more{81} &    -               &   -   \\
&EdgeConnect \cite{naz2019EC}       & 39.99\dg\more{642} & 0.208\dg\more{70} &    -               &   -   \\
&MI-GAN \cite{Sargsyan2023migan}    &      -             &     -             & 27.65\dg\more{239} & 0.358\dg\more{55} \\
&MAT \cite{li2022mat}               &  4.86\dg\less{10}  & 0.125\dg\more{2}  &  9.04\more{11}  & 0.232\more{0} \\
&Latent-C.I. \cite{Chen2024LCI}     &  7.62\more{41}     & 0.153\more{25}    & 19.24\more{136}    & 0.278\more{20}  \\
\midrule
\rowcolor{TabGray} \multicolumn{6}{c}{ \textit{Diffusion-based Methods} } \\
\midrule
\multirow{6}{*}{\rotatebox{90}{\scriptsize\textbf{Academic}}}
&SD \cite{Rombach2022LDM}                & 11.18\more{107} & 0.155\more{27} & 40.24\dg\more{394} & 0.359\dg\more{55}  \\
&PowerPaint \cite{zhuang2023powerpaint}  & 13.43\more{149} & 0.176\more{44} & 38.25\dg\more{369} & 0.409\dg\more{77}  \\
&HD-Painter \cite{manukyan2023hdpainter} & 20.31\more{277}& 0.221\more{81}  & 84.42\more{936} & 0.389\more{68}  \\
&DDNM \cite{wang2022ddnm}                & 14.60\more{171}& 0.680\more{457} & 32.12\more{294} & 0.574\more{148} \\
&RoRem \cite{Li2025RoRem}                & 14.53\more{170}& 0.220\more{80}  & 67.83\more{732} & 0.440\more{90}  \\
&PixelHacker \cite{xu2025pixelhacker}    &  4.75\dg\less{12} & 0.115\dg\less{6} &  6.35\dg\less{22}  & 0.229\dg\less{1} \\
\midrule
\multirow{2}{*}{\rotatebox{90}{\scriptsize\textbf{Indu.}}}
&  SD3.5 Large-Inp. \cite{esser2024SD3} & 11.80\more{119}& 0.134\more{10} &109.42\more{1243} & 0.402\more{74} \\
&  FLUX.1-Fill-Dev \cite{flux2024}      & 10.13\more{88} & 0.141\more{16} & 11.19\more{37}   & 0.268\more{16} \\
\midrule
\midrule
\rowcolor{AblationGreen} & Moebius &  5.39       & 0.122        &  8.15 & 0.231 \\
\bottomrule
\end{tabular}
}
\label{tab:total_portrait}
\vspace{-2em}
\end{wraptable}

\paragraph{\underline{Performance on Portrait Scenes.}}
The capacity of Moebius is further validated in portrait inpainting tasks. As reported in Tab.~\ref{tab:total_portrait}, Moebius achieves 5.39 FID and 0.122 LPIPS on CelebA-HQ. This performance matches the highly specialized MAT \cite{li2022mat} and drastically surpasses all other diffusion models, second only to its teacher. On FFHQ, it obtains consistently leading scores with 8.15 FID and 0.231 LPIPS. Remarkably, Moebius completely eclipses the 10B-level industrial models in this domain, yielding improvements of 37\%–1243\% in FID. The qualitative comparisons in Fig.~\ref{fig:qualitative_comparison} (Right) confirm that Moebius flawlessly reconstructs intricate facial semantics—such as precise eye alignment and skin texture—where massive generalist models often produce structural confusion or blurring.

\newpage
\vspace{1.0em}
\subsection{User Study}\label{subsec:user} 
\vspace{-0.2em}

\begin{figure}[t]
\centering
\includegraphics[width=1.0\linewidth]{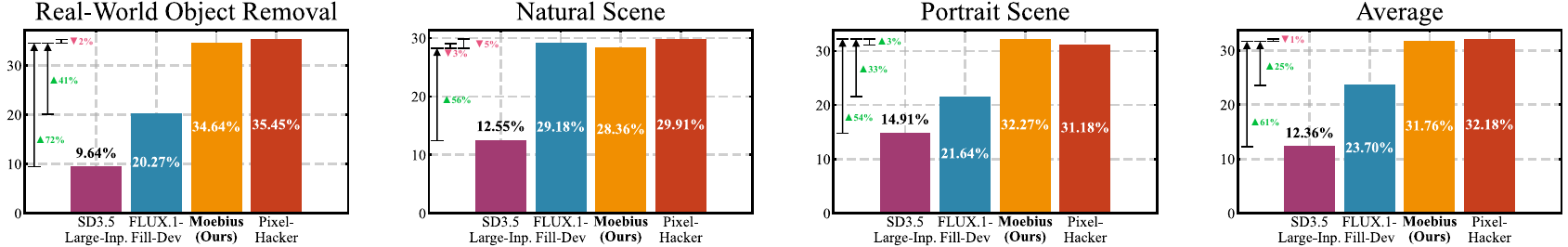}
\caption{
\textbf{User study of Moebius (0.22B) against teacher and 10B-level generalist models across various scenes.}
Moebius matches its teacher's performance and significantly outperforms massive generalists, excelling particularly in portrait scene.
}
\label{fig:user_study}
\vspace{-1em}
\end{figure}

We conducted a double-blind human preference study to assess perceptual quality.
50 cases per scenario were randomly sampled from natural, portrait, and real-world sets. 22 participants (including experts and general users) performed forced-choice tests to select the best result based on global coherence and visual fidelity.
As shown in Fig.~\ref{fig:user_study}, Moebius (31.76\% average) bridges the 50$\times$ parameter gap, matching the teacher PixelHacker (32.18\%) and significantly surpassing 10B-level industrial systems (FLUX.1-Fill-Dev at 23.70\% and SD3.5 Large-Inp. at 12.36\%). Notably, Moebius achieves the highest preference (32.27\%) in portrait scenes, suggesting that for highly structured tasks demanding strict spatial and semantic fidelity, our meticulously distilled specialist captures fine-grained structural nuances more effectively than zero-shot massive generalists.

\vspace{-0.8em}
\subsection{Ablation Study}\label{subsec:abla}

\paragraph{\underline{Further Analysis of Architectural Synergy.}}
Building upon the analysis in Sec.~\ref{sec:architecture_evolution}, Exps \textcircled{\scriptsize 11}-\textcircled{\scriptsize 15} (Tab.~\ref{tab:rebuttal_ablation}) further explore the impact of individual components under our distillation framework. Comparing them against Exp \textcircled{\scriptsize 11} reveals that distilling isolated lightweight operators fails to achieve an optimal quality-efficiency balance. Only the holistic integration of all structural modifications (Exp \textcircled{\scriptsize \textbf{9}}) unlocks the optimal efficiency front, reconfirming that extreme compactness demands rigorous architectural synergy over naive module substitution.

\begin{figure}[t]
\centering
    \begin{minipage}[t]{0.48\textwidth}
        \vspace{0pt}
        \vspace{-1em}
        \centering
        \captionof{table}{
            \textbf{Ablation of optimization objectives in our adaptive multi-granularity distillation strategy.} 
            Evaluated under the identical setup as Tab.~\ref{tab:rebuttal_ablation}, this analysis validates the incremental contribution of each loss.
        }
        \vspace{0.75em}
        \label{tab: abla_distill}
        \resizebox{1.0\linewidth}{!}{
        \begin{tabular}{cccc|cc}
        \toprule
        $\mathcal{L}_\text{C\_KD}$ & 
        $\mathcal{L}_\text{F\_KD}$ & 
        $\mathcal{L}_\text{task}$  &
        $\mathcal{L}_\text{perceptual}$ 
        & $\mathbf{FID}\downarrow$  
        & $\mathbf{LPIPS}\downarrow$  \\
        \midrule
        \cm&   &   &   & 74.20\more{181} & 0.367\more{42} \\
        \cm&\cm&   &   & 36.17\more{37}  & 0.291\more{13} \\
        \cm&\cm&\cm&   & 32.59\more{23}  & 0.273\more{6}  \\ 
        \rowcolor{AblationGreen} \cm&\cm&\cm&\cm& 26.43 & 0.258 \\
        \bottomrule
        \end{tabular}
        }
    \end{minipage}
    \hfill
    \begin{minipage}[t]{0.48\textwidth}
        \vspace{0pt}
        \centering
        \includegraphics[width=1.0\linewidth]{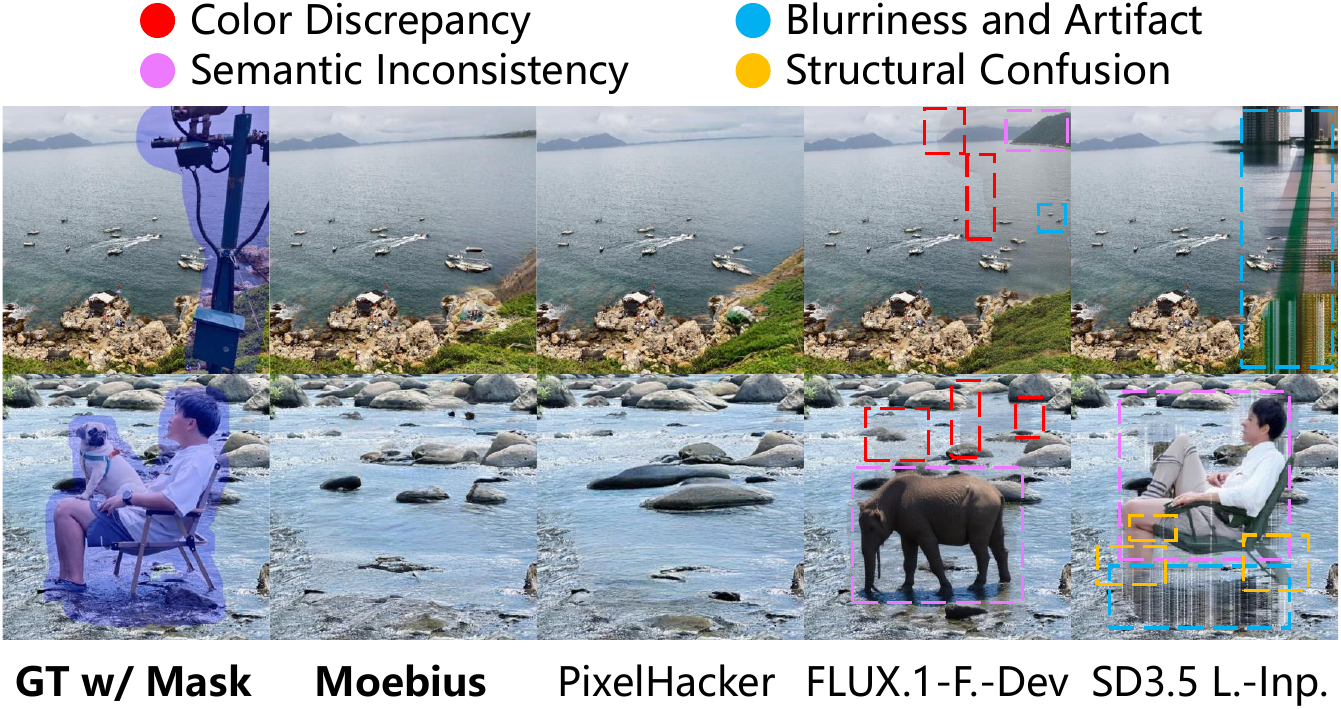}
        \vspace{-1.7em}
        \captionof{figure}{
            \textbf{Real-World Object Removal.} 
            Moebius handles realistic masks with superior consistency compared to baselines.
        }
        \label{fig:removal}
    \end{minipage}
    \vspace{-1em}
\end{figure}

\paragraph{\underline{Dissection of Distillation Objectives.}}
Furthermore, Tab.~\ref{tab: abla_distill} dissects our optimization objectives. Relying solely on the coarse-grained loss ($\mathcal{L}_\text{C\_KD}$) yields a degraded FID of 74.20 for such a compressed network. Progressively integrating fine-grained distillation ($\mathcal{L}_\text{F\_KD}$, $\mathcal{L}_\text{task}$) and the latent perceptual constraint ($\mathcal{L}_\text{perceptual}$) systematically restores the generation quality to 26.43. This confirms that our strictly latent-space multi-granularity optimization is the crucial catalyst for unlocking the full representational capacity of $L\lambda MI$.

\vspace{-0.5em}
\subsection{Real-World Removal Application}\label{subsec:real_world}
To demonstrate the practical deployment potential of Moebius beyond academic benchmarks, we provide qualitative evaluations on real-world object removal tasks, which feature complex contexts and irregular user-drawn masks. As shown in Fig.~\ref{fig:removal}, FLUX.1-Fill-Dev and SD3.5 Large-Inpainting frequently introduce semantic inconsistencies, color discrepancies, and structural confusion (e.g., leaving artifacts when removing the electrical pole, or failing to reconstruct the river after removing the subjects). In stark contrast, Moebius robustly comprehends the global scene context and seamlessly inpaints the missing background, delivering flawless object removal that matches the visual fidelity of PixelHacker.

\vspace{-0.5em}
\section{Conclusion}\label{sec:conclusion}
\vspace{-0.5em}

In this paper, we propose Moebius, a highly efficient specialist that sets a new standard for high-fidelity image inpainting. To conquer the representation bottleneck of extreme structural compression, we synergistically pair our proposed $L\lambda MI$ blocks with an adaptive multi-granularity latent distillation strategy. This optimal synergy empowers our 0.22B-parameter network to rival the generation quality of 10B-level industrial generalists (e.g., FLUX.1-Fill-Dev), while delivering a $>15\times$ total inference acceleration. Ultimately, Moebius proves that extreme compactness can successfully bridge the massive scale gap, advancing resource-constrained deployment.

%
%
\newpage 
\bibliographystyle{splncs04}
\bibliography{main}


\newpage
\section*{Supplementary Materials of Moebius}

\subsection*{Overview}
In this supplementary material, we provide extensive qualitative results and further empirical analyses to reinforce the findings presented in the main manuscript. The contents are organized as follows:
\begin{itemize}
    \item \textbf{Additional Showcases on Natural and Portrait Scenes:} We present large-scale qualitative comparisons across natural (Places2 \cite{zhou2017places}) and portrait (CelebA-HQ \cite{karras2018celebahq}, FFHQ \cite{karras2018stylegan_ffhq}) scenes to demonstrate Moebius's superior fidelity against both 1B-level specialists and 10B-level generalist models.
    \item \textbf{Comparison with Commercial Systems:} We provide a qualitative study comparing Moebius with prominent commercial image editing systems.
    \item \textbf{Failure Case Analysis:} We objectively discuss limitations of Moebius~in~extreme scenarios, reflecting the inherent trade-offs of structural compression.
    \item \textbf{Ablation of Classifier-Free Guidance (CFG):} We provide additional analysis on CFG \cite{ho2022CFG} scale to validate the optimal configuration of Moebius.
    \item \textbf{Evaluation of Out-of-Distribution (OOD) Performance:} We conduct OOD evaluation on both natural and portrait scenes, verifying the strong generalizability and zero-shot capability of Moebius.
\end{itemize}

\subsection*{Additional Showcases on Natural Scenes (Places2)} \label{sec:sup_places}
\begin{figure*}[!ht]
\centering
\includegraphics[width=0.98\linewidth]{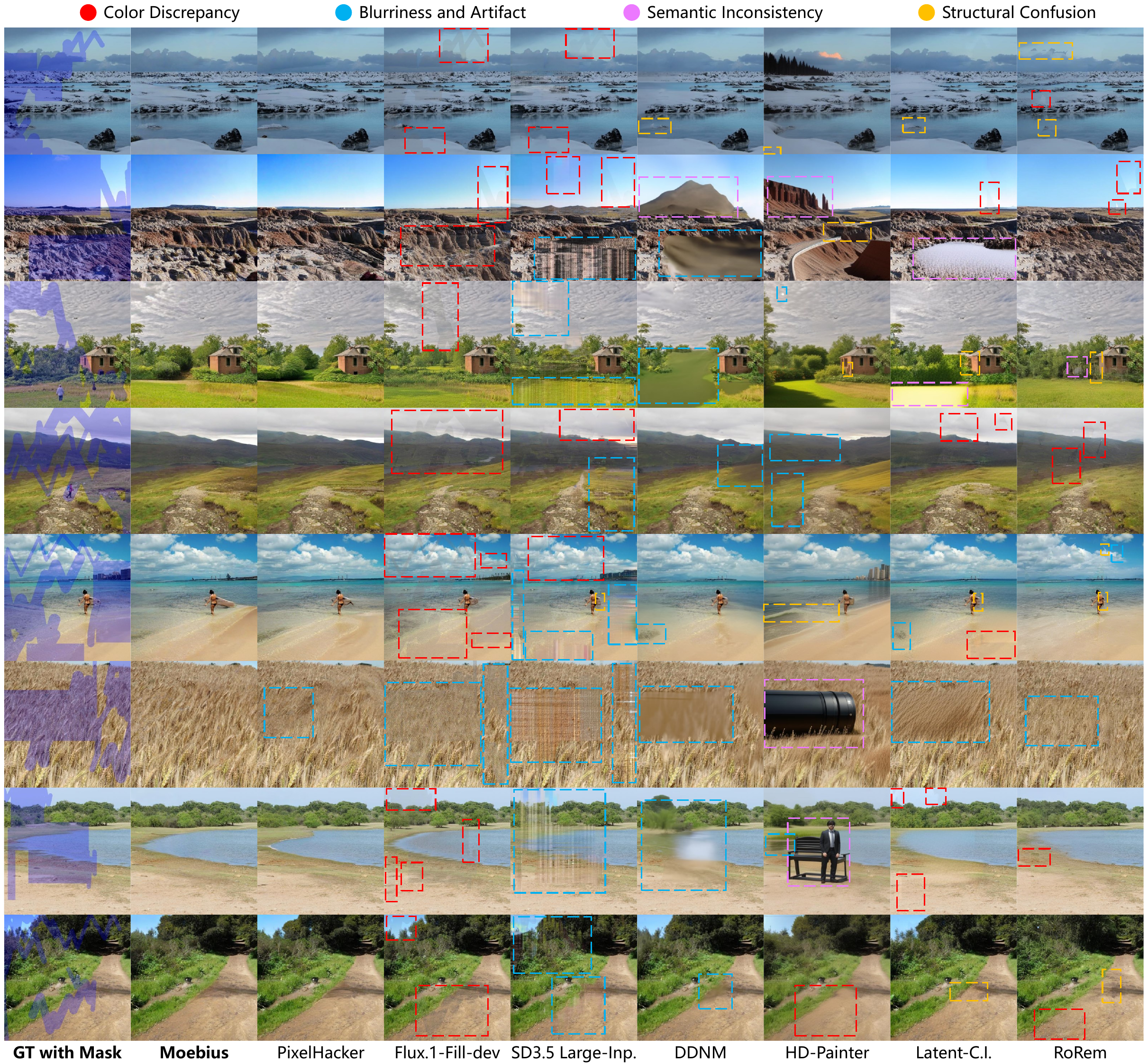}
\caption{\textbf{More qualitative comparison on natural scenes (Places2).} 
Even with significantly lower parameter counts, Moebius still achieves better contextual consistency across diverse natural scenes compared with other methods, including avoiding structural confusion in rocky terrains (row 2), large-scale blurriness or artifacts in rice fields (row 6), and inconsistent generation in beach and land regions (rows 7 and 8).}
\label{fig:sup_showcase_places}
\vspace{-1em}
\end{figure*}

We provide an additional extensive qualitative comparison on the Places2~\cite{zhou2017places} dataset. As illustrated in Fig.~\ref{fig:sup_showcase_places}, our showcases cover diverse natural contexts, including complex architectures, and vegetation. Compared to the 10B-level industrial generalists (FLUX.1-Fill-Dev \cite{flux2024} and SD3.5 Large-Inp. \cite{esser2024SD3}), Moebius demonstrates superior contextual consistency. While massive generalist models occasionally suffer from semantic shift or over-generation (hallucinating objects irrelevant to the background), Moebius faithfully preserves the structural continuity of the original scene. Furthermore, despite its extreme 0.22B parameter count, Moebius inherits the precise texture synthesis capabilities of its teacher (PixelHacker \cite{xu2025pixelhacker}), effectively bridging the capacity gap through our proposed $L\lambda MI$ blocks and adaptive distillation. This evidence underscores Moebius's effectiveness as a highly optimized specialist for natural scene reconstruction.

\subsection*{Additional Showcases on Portrait Scenes (CelebA-HQ, FFHQ)} \label{sec:sup_portrait}
\begin{figure}[!ht]
\centering
\includegraphics[width=0.98\linewidth]{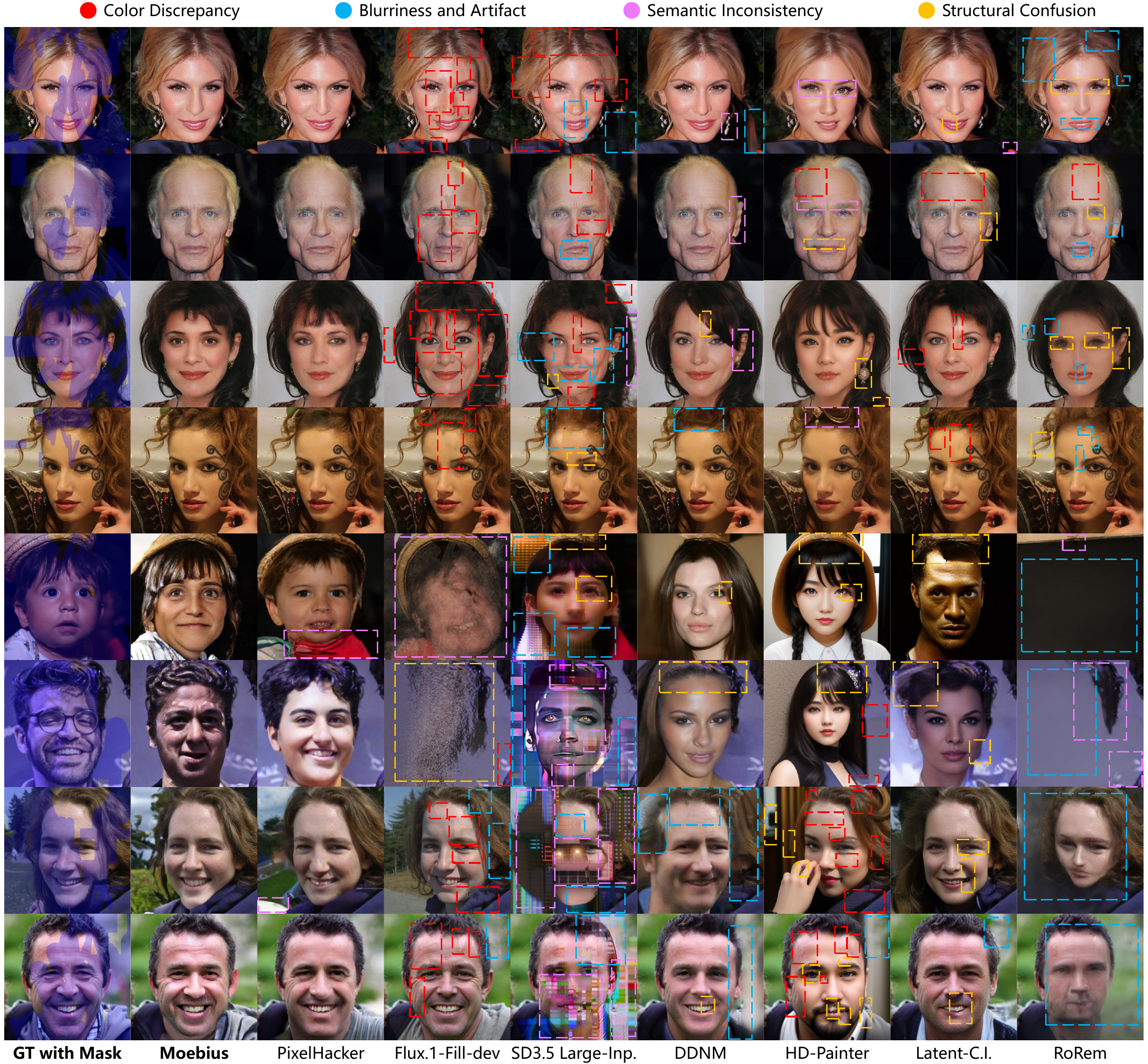}
\caption{
\textbf{More qualitative comparison on portrait scenes (CelebA-HQ and FFHQ).} 
Rows 1–4: CelebA-HQ; rows 5–8: FFHQ. Moebius excels in facial restoration, maintaining superior structural integrity and skin texture fidelity compared to other methods. 
Specifically, Moebius provides sharper and more coherent results, avoiding color discrepancies on face (row 1), reducing blurriness and artifacts at mouth and ear (rows 2 and 3), suppressing semantic inconsistencies in hair (rows 4 and 5), and preventing structural confusion at teeth and background (rows 6, 7 and 8).
}
\label{fig:sup_showcase_portrait}
\vspace{-2.5em}
\end{figure}

As shown in Fig.~\ref{fig:sup_showcase_portrait}, we provide more showcases from CelebA-HQ~\cite{karras2018celebahq} and FFHQ~\cite{karras2018stylegan_ffhq} to evaluate the performance of Moebius in restoring complex facial features. Moebius demonstrates a remarkable ability to capture facial symmetry and intricate skin textures. While 10B-level generalist models (FLUX.1-Fill-Dev \cite{flux2024} and SD3.5 Large-Inp. \cite{esser2024SD3} ) occasionally introduce structural confusion when the mask covers critical facial components, Moebius consistently generates plausible results. This superior performance on portrait tasks even surpasses its teacher model in several instances, which aligns with our user study findings (Sec.~4.3 in the main paper), proving that a task-specific specialist can excel at modeling domain-specific distributions with a fraction of the parameters.

\begin{figure}[!ht]
\centering
\includegraphics[width=0.83\linewidth]{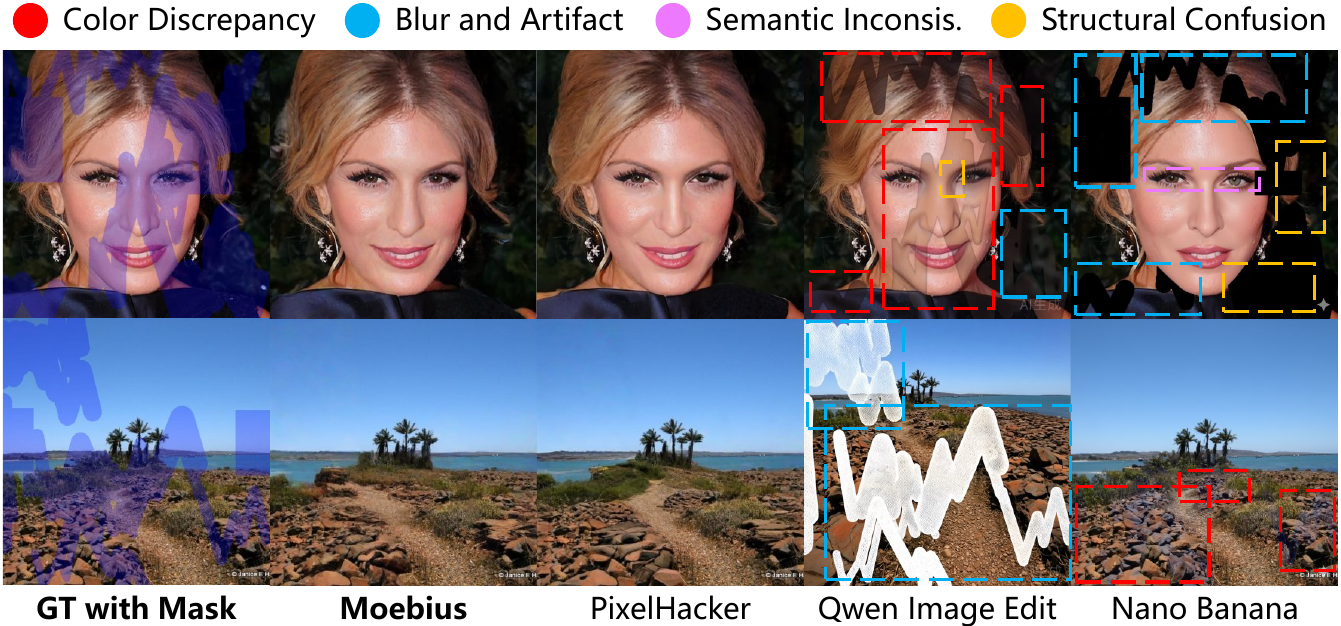}
\vspace{-0.5em}
\caption{
\textbf{Qualitative comparison with commercial edit systems.} 
Despite massive scale gap, Moebius demonstrates visual quality comparable to commercial models (Nano Banana \& Qwen Image Edit), effectively handling complex details restoration.
}
\label{fig:nanobanana_qwen}
\end{figure}

\subsection*{Comparison with Commercial Edit Models} \label{sec:sup_commercial}
We compare Moebius against prominent commercial-grade image editing systems, including Nano Banana \cite{2025nano_banana} and Qwen Image Edit \cite{wu2025qwenimagetechnicalreport}. These models represent the current industry standards for high-end generative editing. As illustrated in Fig.~\ref{fig:nanobanana_qwen}, Moebius demonstrates highly competitive visual fidelity relative to these large-scale commercial systems. Although Nano Banana and Qwen Image Edit are backed by massive computational resources and extreme parameter scales, Moebius, an extremely lightweight 0.22B open-source specialist, faithfully restores missing regions with sharp textures and global coherence. This comparison underscores the profound practical applicability and accessibility of our highly optimized architecture, proving that state-of-the-art generative capabilities can be achieved within a significantly constrained parameter budget.

\vspace{-0.5em}
\subsection*{Failure Case Analysis} \label{sec:sup_failure}
\begin{figure}[htbp]
\centering
\includegraphics[width=1.0\linewidth]{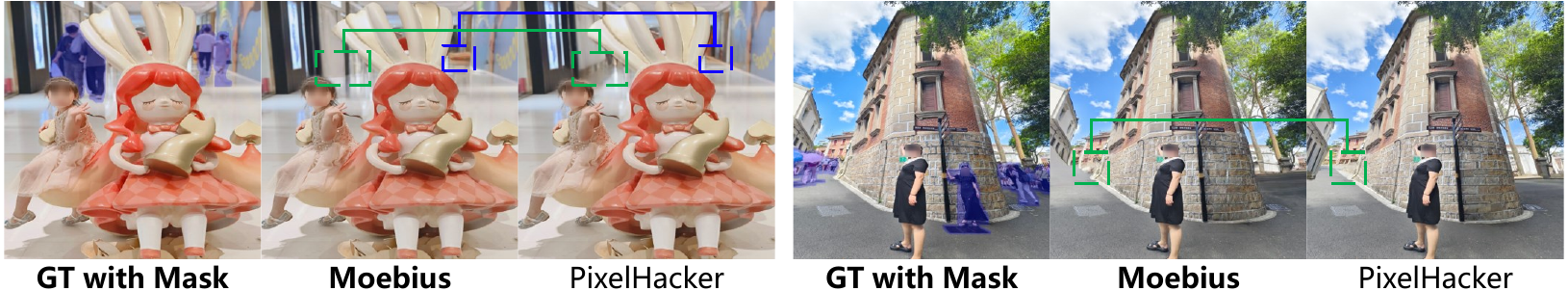}
\vspace{-1.5em}
\caption{\textbf{Failure case analysis.} Compared with its teacher model (PixelHacker), Moebius may exhibit minor detail loss or less plausible textures in extremely tiny background regions when context is limited. These instances illustrate the capacity-efficiency trade-off of our lightweight specialist.}
\label{fig:sup_showcase_badcase}
\end{figure}

\vspace{-2.5em}
Despite its superior performance, Moebius is subject to certain limitations inherent to extreme structural compression. As illustrated in Fig.~\ref{fig:sup_showcase_badcase}, we compare Moebius with its teacher model, PixelHacker \cite{xu2025pixelhacker}, in several challenging scenarios. While Moebius excels in restoring most structured and natural contexts, it occasionally faces challenges in restoring fine-grained geometry within tiny background aesthetics. For instance, in tiny background inpainting with severely limited contextual textures, Moebius may produce slightly less plausible details compared to its 1B-parameter teacher. This minor degradation represents an acceptable compromise within our rigorous efficiency-parameter-performance trade-off, reflecting the inherent capacity limits of a 0.22B-parameter framework when striving for extreme structural compactness.

\begin{table}[!ht]
\caption{
\textbf{Ablation of CFG scales.}
To validate the optimal CFG configuration on natural (Places2) and portrait (CelebA-HQ) scenes, we conducted extensive ablation studies on Places2 (Test) and CelebA-HQ (512) benchmarks. 
Results indicate that the optimal CFG scales for natural and portrait scenes are 2.5 and 2.0 (highlighted in \textbf{\textcolor{AblationGreenWord}{green}}), which serve as our default settings.
}
\vspace{-0.5em}
\centering
\resizebox{1.0\linewidth}{!}{
    \begin{tabular}{
        c|>{\hspace{1pt}}c<{\hspace{1pt}}>{\hspace{1pt}}c<{\hspace{1pt}}>{\hspace{1pt}}>{\columncolor{AblationGreen}}c<{\hspace{1pt}}>{\hspace{1pt}}c<{\hspace{1pt}}>{\hspace{1pt}}c<{\hspace{1pt}}>{\hspace{1pt}}c<{\hspace{1pt}}>{\hspace{1pt}}c<{\hspace{1pt}}
        ||
        c|>{\hspace{1pt}}c<{\hspace{1pt}}>{\hspace{1pt}}c<{\hspace{1pt}}>{\hspace{1pt}}>{\columncolor{AblationGreen}}c<{\hspace{1pt}}>{\hspace{1pt}}c<{\hspace{1pt}}>{\hspace{1pt}}c<{\hspace{1pt}}>{\hspace{1pt}}c<{\hspace{1pt}}>{\hspace{1pt}}c<{\hspace{1pt}}
        }
    \toprule
        $\textbf{\textit{Scale}}$ & & & & & & & & $\textbf{\textit{Scale}}$ & & & & & & & \\
        $\textbf{\textit{(Nature)}}$  
            & \multirow{-2}{*}{1.5}  
            & \multirow{-2}{*}{2.0} 
            & \multirow{-2}{*}{2.5}
            & \multirow{-2}{*}{3.0} 
            & \multirow{-2}{*}{3.5} 
            & \multirow{-2}{*}{4.0} 
            & \multirow{-2}{*}{4.5}
        & $\textbf{\textit{(Portrait)}}$  
            & \multirow{-2}{*}{1.0}  
            & \multirow{-2}{*}{1.5} 
            & \multirow{-2}{*}{2.0}
            & \multirow{-2}{*}{2.5} 
            & \multirow{-2}{*}{3.0} 
            & \multirow{-2}{*}{3.5} 
            & \multirow{-2}{*}{4.0}   \\
    \midrule
        $\mathbf{FID}\downarrow$ & 10.14 & 9.65 & 9.48 & 9.51 & 9.73 & 10.09 & 10.37 
        & $\mathbf{FID}\downarrow$  & 5.54 & 5.42 & 5.39 & 5.49 & 5.61 & 5.74 & 5.91 \\
        $\mathbf{LPIPS}\downarrow$ & 0.209 & 0.208 & 0.207 & 0.208 & 0.209 & 0.210 & 0.212 
        & $\mathbf{LPIPS}\downarrow$  & 0.126 & 0.123 & 0.122 & 0.122 & 0.123 & 0.124 & 0.126 \\
    \bottomrule
    \end{tabular}
}
\label{tab: abla_CFG}
\vspace{-1.5em}
\end{table}

\vspace{-1.5em}
\subsection*{Ablation Study on Classifier-Free Guidance Scale} \label{sec:sup_cfg}
Similar to standard latent diffusion models \cite{Rombach2022LDM,esser2024SD3,xu2025pixelhacker}, we employ Classifier-Free Guidance (CFG) \cite{ho2022CFG} to balance sample quality and diversity during inference. Tab.~\ref{tab: abla_CFG} reports the quantitative performance across varying CFG scales for both natural (Places2) and portrait (CelebA-HQ) scenes. These evaluations are conducted using the 51k-step checkpoint for Places2 and the 60k-step checkpoint for CelebA-HQ, respectively. As shown in the results, the optimal CFG scale that achieves the best trade-off between FID and LPIPS is 2.5 for natural scenes and 2.0 for portrait scenes. Consequently, we adopt these specific scales as the default inference settings for all corresponding experiments in our framework.

\subsection*{Evaluation of Out-of-Distribution (OOD) Performance} \label{sec:ood}
We conduct an OOD evaluation, as shown in Tab.~\ref{tab:abla_ood}.
Moebius maintains strong results on both OOD natural and OOD portrait scenes, demonstrating its excellent generalizability and zero-shot capability.

\begin{table}[htbp]
\caption{
\textbf{OOD evaluation for verifying generalization \& zero-shot capability.} 
We sampled 10k images from LVIS\cite{gupta2019lvis} as OOD natural dataset and 3k images from the wiki subset of DeepFakeFace\cite{song2023deepfakeface} as OOD portrait dataset. Mask settings follow the evaluation protocol of Places2 (Test)/CelebA-HQ (512) in main paper. All academic methods, including Moebius, are evaluated using available Places2/CelebA-HQ weights.
The strong results demonstrate Moebius' excellent generalizability.
}
\centering
\resizebox{1.0\linewidth}{!}{
    \begin{tabular}{>{\hspace{2pt}}c<{\hspace{2pt}}|l|*{2}{w{c}{1.8cm}}|*{2}{w{c}{2.5cm}}}
\toprule
    \multicolumn{2}{c|}{\multirow{2}{*}{\textbf{Method}}}
    & \multicolumn{2}{c|}{\textbf{OOD Natural (LVIS)}}
    & \multicolumn{2}{c }{\textbf{OOD Portrait (DeepFakeFace)}}\\
    \multicolumn{2}{c|}{}
    &  $\mathbf{FID}\downarrow$ & $\mathbf{LPIPS}\downarrow$
    &  $\mathbf{FID}\downarrow$ & $\mathbf{LPIPS}\downarrow$ \\
    \midrule
\multirow{4}{*}{\rotatebox{90}{\scriptsize\textbf{Academic}}}
& MAT\cite{li2022mat} & 18.08\more{2} & 0.312\more{1} & 25.60\more{67} & 0.203\more{17} \\
& MI-GAN\cite{Sargsyan2023migan} & 25.88\more{45} & 0.347\more{12} & - & - \\
& Latent-C.I.\cite{Chen2024LCI} & 31.94\more{79} & 0.399\more{29} & 43.07\more{181} & 0.327\more{89} \\
& PixelHacker\cite{xu2025pixelhacker} & 13.84\less{22} & 0.305\less{1} & 15.50\more{1} & 0.172\less{1} \\
    \midrule
\multirow{2}{*}{\rotatebox{90}{\scriptsize\textbf{Indu.}}}
& SD3.5 Large-Inp.\cite{esser2024SD3} & 114.21\more{541} & 0.601\more{94} & 81.91\more{435} & 0.372\more{115} \\
& FLUX.1-Fill-dev\cite{flux2024}  & 14.52\less{18}   & 0.339\more{10} & 12.24\less{20} & 0.174\more{1} \\
    \midrule
\rowcolor{AblationGreen} & Moebius & 17.81 & 0.309 & 15.32 & 0.173 \\
\bottomrule
    \end{tabular}
}
\label{tab:abla_ood}
\end{table}

\end{document}